\pdfoutput=1

\documentclass[11pt]{article}

\usepackage[]{EMNLP2022}

\usepackage{times}
\usepackage{latexsym}

\usepackage[T1]{fontenc}

\usepackage[utf8]{inputenc}

\usepackage{microtype}

\usepackage{inconsolata}

\usepackage{graphicx}
\usepackage{caption}
\usepackage{subcaption}
\usepackage{amsmath}
\usepackage{amssymb}
\usepackage[export]{adjustbox}

\usepackage{cleveref}
\crefname{section}{§}{§§}
\Crefname{section}{§}{§§}
\usepackage{xcolor}

\definecolor{kmy-color}{rgb}{0.28, 0.58, 0.28}

\newcommand{\bertbase}{\texttt{bert-base-uncased}}
\newcommand{\adapteremoji}{
    \includegraphics[scale=0.7]{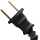}
}

\newcommand{\fireemoji}{
    \includegraphics[]{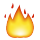}
}

\newcommand{\ft}{fine-tune}
\newcommand{\fting}{fine-tuning}
\newcommand{\systemname}{\textsc{UDApter}}
\newcommand{\dann}{\textsc{dann}}
\newcommand{\dsn}{\textsc{dsn}}
\newcommand{\dsnadapter}{\dsn{}-\adapteremoji}
\newcommand{\dannadapter}{\dann{}-\adapteremoji}
\newcommand{\dannadaptermc}{\dann{}-\adapteremoji-\textsc{mc}}
\newcommand{\taskadapter}{\textsc{task}-\adapteremoji{}}
\newcommand{\ourmethod}{\textsc{ts-dt}-\adapteremoji{}}
\newcommand{\uda}{\textsc{uda}}
\newcommand{\jointdt}{\textsc{joint-dt}-\adapteremoji{}}

\newcommand{\apparel}{\textsc{a}}
\newcommand{\baby}{\textsc{ba}}
\newcommand{\books}{\textsc{bo}}
\newcommand{\cameraphoto}{\textsc{c}}
\newcommand{\moviereviews}{\textsc{mr}}

\newcommand{\travel}{\textsc{tr}}
\newcommand{\telephone}{\textsc{te}}
\newcommand{\fiction}{\textsc{f}}
\newcommand{\government}{\textsc{g}}
\newcommand{\slate}{\textsc{s}}

\newcommand{\amazondataset}{\textsc{amazon}}
\newcommand{\mnlidataset}{\textsc{mnli}}

\newcommand{\declarelogo}[0]{\includegraphics[height=.02\textwidth]{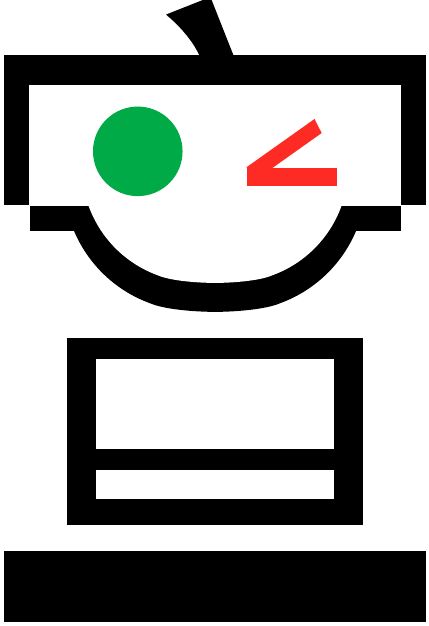}}

\title{\textsc{UDApter} - Efficient Domain Adaptation Using Adapters}

\author{Bhavitvya Malik$^{*\ \alpha}$,  Abhinav Ramesh Kashyap \thanks{\ \ The first two authors contributed equally. } \ $^{\beta \gamma}$,\\ \textbf{Min-Yen Kan$^{\beta}$, Soujanya Poria}$^{\declarelogo}$\\
$^{\alpha}$ The University of Edinburgh, Edinburgh\\
$^{\beta}$ National University of Singapore, Singapore\\
$^{\gamma}$ ASUS Intelligent Cloud Services (AICS), Singapore\\
$^{\declarelogo}$ DeCLaRe Lab, Singapore University of Technology and Design, Singapore \\
\fontsize{10}{10}\texttt{b.malik-1@sms.ed.ac.uk,}\ \fontsize{10}{10}\texttt{abhinav\_kashyap@asus.com,}\ \fontsize{10}{10}\texttt{kanmy@comp.nus.edu.sg,}\\ 
\fontsize{10}{10}\texttt{sporia@sutd.edu.sg}\\ 
}

\begin{document}
\maketitle
\begin{abstract}

We propose two methods to make unsupervised domain adaptation (\uda{}) more parameter efficient using adapters, small bottleneck layers interspersed with every layer of the large-scale pre-trained language model (PLM). The first method deconstructs \uda{} into a two-step process: first by adding a \textit{domain adapter} to learn domain-invariant information and then by adding a \textit{task adapter} that uses domain-invariant information to learn task representations in the source domain. The second method jointly learns a supervised classifier while reducing the divergence measure. 
Compared to strong baselines, our simple methods perform well in natural language inference (\mnlidataset) and the cross-domain sentiment classification task. We even outperform unsupervised domain adaptation methods such as DANN \cite{dann} and DSN \cite{dsn} in sentiment classification, and we are within 0.85\% F1 for natural language inference task, by \fting{} only a fraction of the full model parameters. We release our code at \textit{https://github.com/declare-lab/domadapter}.

\end{abstract}

\begin{figure*}[t!]
    \centering
    \subfloat[ \label{fig:domain-adapter}]{\includegraphics[scale=0.5]{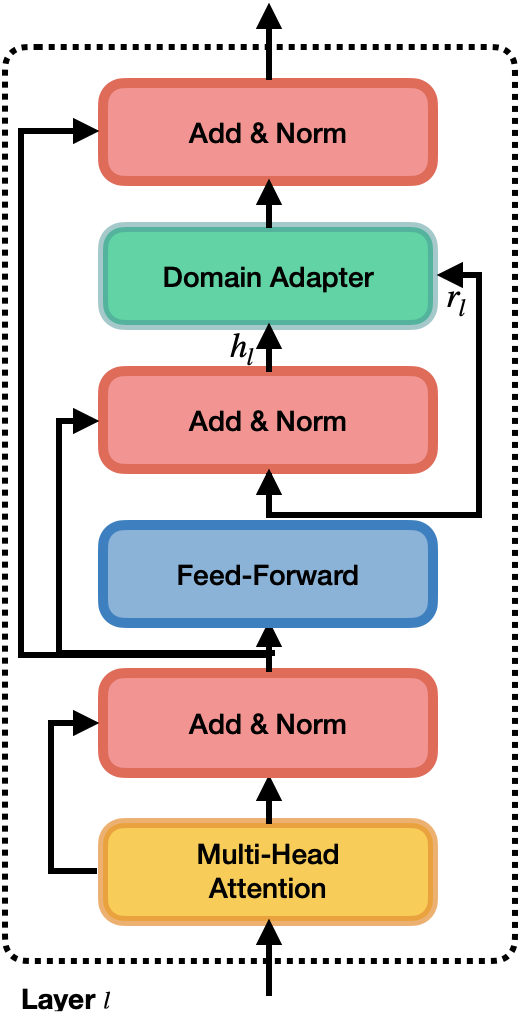}} 
    \hspace{0.6cm}
    \subfloat[\label{fig:task-adapter}]{\includegraphics[scale=0.5]{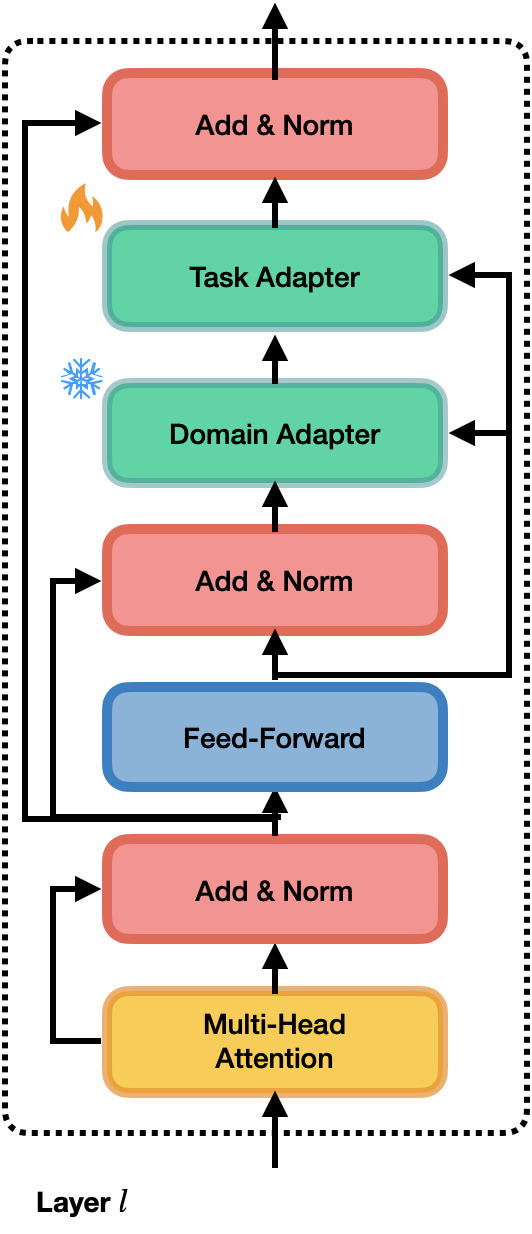}} 
    \hspace{0.6cm}
    \subfloat[\label{fig:domain_and_task_adapter}]{\includegraphics[scale=0.5]{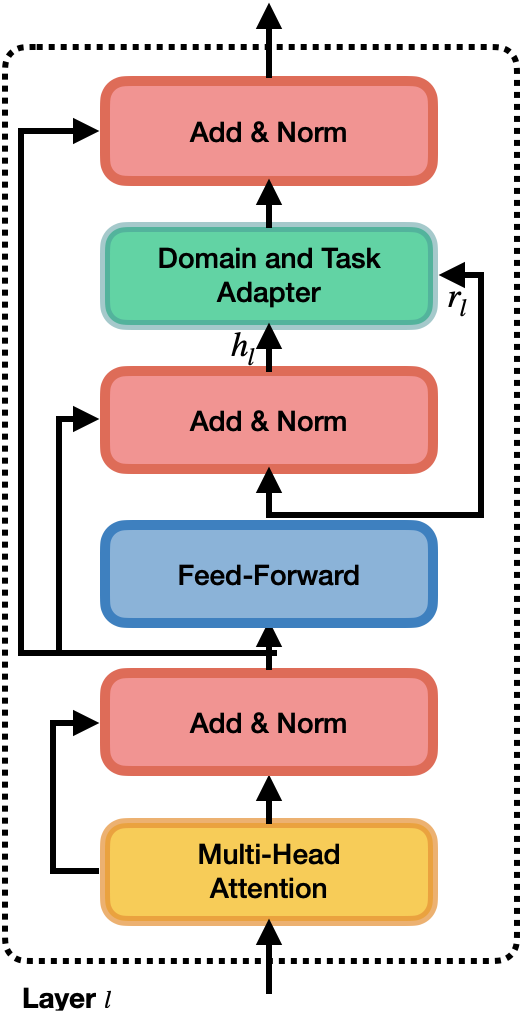}} 
    \caption{\systemname{} for a transformer layer $l$ uses principles from unsupervised domain adaptation to make domain adaptation more parameter efficient. (a) The first method \ourmethod{} trains a Domain Adapter that reduces the marginal distribution between the domains (b) The task adapter is stacked on top of the domain adapter. and trained on an end task like sentiment analysis or natural language inference. The domain adapter is frozen during training. (c) The second method \jointdt{} reduces the domain divergence and the task loss jointly. }
    \label{fig:architecture}
\end{figure*}
\section{Introduction}

Fine-tuning pretrained language models (PLM) is the predominant method for improving NLP tasks such as sentiment analysis, natural language inference, and other language understanding tasks \cite{wang-etal-2018-glue}.  However, \fting{} forces us to modify all the parameters of the model and store one copy of the model for one task. Given the large size of current PLMs, this can be expensive. Furthermore, \fting{} needs large-scale data to be effective and is unstable when using different seeds \cite{han-etal-2021-robust}. 

A new approach to alleviate this is parameter-efficient \fting{} -- freezing the PLM parameters and \fting{} only a small fraction of the parameters. Fine-tuning with adapters \cite{pmlr-v97-houlsby19a} is one of these methods in which small additional layers are tuned within each PLM layer. Fine-tuning with adapters has many advantages: performance comparable to full \fting{} \cite{towards-a-unified-view-of-parameter-efficient-transfer-learning}, and robustness to different seeds and adversarial examples \cite{han-etal-2021-robust}.

Unsupervised domain adaptation (\uda{}) aims to adapt models to new domains and considers situations where labeled data are available only in the source domain and unlabeled data are available in the target domain. \uda{} methods in general have two components: The first reduces the divergence between the source and target domains, and the second reduces the loss corresponding to a particular task \cite{ramesh-kashyap-etal-2021-domain}. However, they \ft{} a large number of parameters and are susceptible to catastrophic forgetting. Adapters \cite{pmlr-v97-houlsby19a} can help solve these problems. However, the benefits of using adapters \fting{} for domain adaptation have been mostly overlooked.  \textit{How well can adapter fine-tuning perform across different domains. Can we make domain adaptation more efficient?} In this work, we answer these questions and propose models to perform domain adaptation using adapters.

Adapters are known to perform well in low-resource scenarios where a small amount of supervised data is available in a new domain or language \cite{he-etal-2021-effectiveness, pfeiffer-etal-2020-mad}. In this work, using the principles of \uda{}, we propose to make domain adaptation more effective using unsupervised data from the target domain. We introduce two methods that we collectively call the \textbf{U} nsupervised \textbf{D} omain \textbf{A} daptation method using ada\textbf{pters} (\systemname{}). The first method is a two-step process: First, we learn \textit{domain adapters} -- where we use a divergence measure to bring two probabilistic distributions closer together. This helps us to learn representations that are independent of the domain from which they come. Second, we use the domain-invariant information learned as input to another task adapter that learns to perform an NLP task using labeled data from the source domain. We combine the two adapters by stacking them. The second method adds a single adapter without stacking, where we simultaneously reduce the divergence between domains and learn the task in the source domain. 

Domain Adversarial Neural Networks (\dann{}) and Domain Separation Networks (\dsn{}) are the most common methods for unsupervised domain adaptation in NLP \cite{ramesh-kashyap-etal-2021-domain}. We compare our proposed methods with these strong baselines that fine-tune all model parameters, on Amazon \cite{blitzer-etal-2007-biographies} and the MNLI dataset \cite{williams-etal-2018-broad} consisting of five domains each. \systemname{} performs better than all baselines. It achieves competitive performance compared to UDA methods by fine-tuning only a fraction of the parameters. In an era where large resources are spent to further pretrain language models on large amounts of unsupervised data to achieve domain adaptation \cite{gururangan-etal-2020-dont}, it is necessary to provide cheaper, faster solutions.

\section{Method}
\label{sec:method}


\paragraph{Setup.} We consider an NLP task (sentiment analysis) consisting of data $\mathcal{X}$ and labels $\mathcal{Y}$ (positive, negative). There exist two different distributions, called the source domain $\mathcal{D_S}$ and the target domain $\mathcal{D_T}$ over $\mathcal{X} \times \mathcal{Y}$. Unsupervised domain adaptation (\uda{}) consists of a model $\mathcal{C}$ that receives labeled input samples $\mathcal{X_S}: (x_s, y_s)_{s=1}^{n_s} \sim \mathcal{D}_\mathcal{S}$ and unlabeled input $\mathcal{X_T}: (x_t)_{t=1}^{n_t} \sim \mathcal{D}_\mathcal{T}$. The goal of \uda{} is to learn a model $\mathcal{C}$ such that we perform well in the NLP task for the target domain $\mathcal{D_T}$.

The popular method in \uda{} is to learn representations that are invariant in the input domain and still have sufficient power to perform well in the source domain \cite{dann, dsn}. 
Then, according to the theory of domain divergence \cite{ben2010theory} shows that the error in the target domain is bounded by the error in the source domain and the divergence. The unsupervised domain adaptation method thus consists of two components: the reduction of the divergence measure and a classifier for the source domain. A new classifier must be learned for every pair of source-target domains, and the method fine-tunes a large number of parameters.  

\systemname{} makes unsupervised domain adaptation more parameter efficient (cf. \Cref{sec:two-step-method}, \Cref{sec:joint-method}) using adapters. We follow the framework proposed by \citet{pmlr-v97-houlsby19a} where small bottleneck layers are added to the transformer layers, \fting{} only the adapter parameters while keeping the other parameters frozen, and propose the following. 

\subsection{Two-Step Domain and Task Adapters}
\label{sec:two-step-method}
\paragraph{Domain Adapters.} To learn domain-invariant representations, we first train a domain adapter. The adapter architecture follows the work of \citet{pfeiffer-etal-2021-adapterfusion}, which consists of a simple down-projection followed by an up-projection. In a transformer layer $l$, let $h_l$ be the hidden representation of the layer \textbf{Add \& Norm}  and let $r_l$ be the representation of the layer \textbf{Feed-Forward} (\Cref{fig:domain-adapter}), then the adapter makes the following transformation and calculates a new hidden representation.

\begin{equation}
    dom_{l} = W_{up} \cdot f(W_{down} \cdot  h_l) + r_l
\end{equation}

\noindent where $f$ is a nonlinear function (e.g., \textsc{ReLU}), $W_{down} \in \mathbb{R}^{h \times d}$ projects the hidden representations down to a lower dimension, $W_{up} \in \mathbb{R}^{d \times h}$ projects them back to a higher dimension, and $d \ll h$. We pass a sample from the source domain $(x_s^{src}) \sim \mathcal{D_S}$ and one from the target $(x_t^{trg}) \sim \mathcal{D_T}$ through the adapters in layer $l$ and obtain their representations $h^{src}_l$ and $h^{trg}_l$, respectively. We then reduce the divergence between these representations.

\begin{equation}
    \Delta_l = div(dom_l^{src}, dom_l^{trg})
\end{equation}

Here $div(\cdot)$ is the divergence function such as the correlation alignment (CORAL) \cite{coral}, the central moment discrepancy (CMD) \cite{cmd} or the multi-kernel maximum mean discrepancy (MK-MMD) \cite{mmd, dsn}. In this work, we use MK-MMD for all of our experiments, since it performed best\footnote{CMD and CORAL also perform similarly to MK-MMD}.  Similar ideas are used to adapt representations in computer vision models \cite{deepadaptationnetworks, deepcoral}.
The final divergence loss considers all $L$ layers.

\begin{equation}
    \mathcal{L}_{div} = \sum_{l=1}^{L} \Delta_l
\end{equation}

\paragraph{Task Adapters.} Task adapters are stacked with frozen domain adapters. We pass the representations $dom_l$ from the previous step and the supervised data from the source domain $(x_{s}^{src}, y_s^{src}) \sim \mathcal{D_S}$. Task adapters have the same architecture as domain adapters and perform the following:

\begin{equation}
    task_l = W_{up} \cdot f(W_{down} \cdot dom_l^{src}) + r_l
\end{equation}

The goal of these task adapters is to learn representations that are task-specific. Only task adapters are updated when training on the end task (sentiment classification, natural language inference) and all other parameters, including domain adapters, are frozen. Regular cross-entropy loss is reduced during training of task adapters:

\begin{equation}
    \mathcal{L}_{task} = softmax\_ce(W_{task} \cdot h_L)
\end{equation}

$h_L$ is the hidden representations of the last layer of the transformer, $W_{task} \in \mathbb{R}^{h * |\mathcal{Y}|}$ where $|\mathcal{Y}|$ is the number of classes, and $softmax\_ce$ is the softmax followed by cross-entropy.  This two-step process deconstructs \uda{} methods with a domain adapter and a task adapter. This affords composability, where task adapters can be reused for different pairs of domains (\Cref{sec:further-analysis}). However, domain and task representations can be learned jointly, as explored in the next section.

\paragraph{Training Process.} Given a source-target domain adaptation scenario, we first train the domain adapter and save their weights. We then stack the task adapter with the domain adapter, which is trained using the supervised data from the source domain. When training the task adapter, the domain adapter is frozen. During inference, we stack the domain and task adapter.

\subsection{Joint Domain Task Adapters}
\label{sec:joint-method}
This method adds a single adapter that performs the reduction of the divergence measure and learns task representations jointly. For a given supervised sample from the source domain $(x_s^{src}, y_s^{src}) \sim \mathcal{D_S}$ and an unsupervised sample $(x_t^{trg}) \sim \mathcal{D_T}$, let $h_l^{src}, h_l^{trg}$ be the hidden representations of the adapters for $x_s^{src}$ and $x_t^{trg}$ for layer $l$. We reduce the following joint loss:

\begin{equation}
 \mathcal{L} = \lambda \cdot \mathcal{L}_{task} + (1-\lambda) \cdot \mathcal{L}_{div}
\end{equation}

Here $\mathcal{L}_{task}$ is the task loss on the source domain supervised samples, $\lambda$ is the adaptation factor.

Reducing divergence along with cross-entropy loss beyond a certain point makes training unstable and does not contribute to increased performance. Following \cite{dann} we suppress the noisy signal from the divergence function as training progresses and gradually change $\lambda$ from 0 to 1 to reduce the contribution of divergence loss using the following schedule ($\gamma=10$ for all of our experiments):

\begin{equation}
    \lambda = \frac{2}{1+\exp{(-\gamma \cdot p)}} - 1 
\end{equation}

Similar methods have been proposed to adapt models to other domains by \citet{deepadaptationnetworks} and \citet{wu-etal-2022-learning}. Compared to the two-step process introduced earlier (\Cref{sec:joint-method}), we need to properly control the losses to obtain optimal results and also this method does not offer composability (\Cref{sec:further-analysis}).

\begin{table}[t!]
\centering
 \begin{tabular}{c@{\hskip 0.2in}r@{\hskip 0.2in}r@{\hskip 0.2in}r} 
 \hline
 Dataset & Train & Dev & Test \\ [0.5ex] 
 \hline
 \mnlidataset{} & 69,600 & 7,730 & 1,940 \\ 
 \amazondataset{} & 1,440 & 160 & 400 \\
 \hline
 \end{tabular}
 \caption{Dataset statistics, showing number of train, dev, and test instances per domain.}
 \label{tab:dataset}
\end{table}

\section{Experiments}

\renewcommand{\arraystretch}{1.2}
\begin{table*}[t!]
\centering \footnotesize 
\resizebox{\linewidth}{!}{%
 \begin{tabular}{l|c|@{\hskip 0.4in}cc|ccccc} 
 \hline
 & \colorbox{red!30}{\textbf{Fully Supervised}} & \multicolumn{2}{@{\hskip -0.3in}c@{\hskip 0.1in}|}{\colorbox{orange}{\textbf{Unsupervised Domain Adaptation}}} & \multicolumn{5}{c}{\colorbox{blue!25}{\textbf{Adapter Based}}} \\ 
 Src \textrightarrow Trg & \fireemoji & \dann{} & \dsn{} & \dannadapter{} & \dannadaptermc{} & \taskadapter{} & \ourmethod{} & \jointdt{} \\ [0.5ex]
 \hline
    \apparel{} \textrightarrow{} \baby{} & 87.52 \scriptsize(1.96) & 85.57 \scriptsize(3.72) & 89.90 \scriptsize(0.26) & 86.46 \scriptsize(0.26) & \textbf{88.74 \scriptsize (0.64)} & 87.03 \scriptsize(0.26) & 88.24 \scriptsize(0.76) & \textbf{88.74 \scriptsize (0.13)} \\
    \apparel{} \textrightarrow{} \books{} & 86.67 \scriptsize(1.06) & 36.48 \scriptsize(0.45) & 84.47 \scriptsize(0.99) & 78.41 \scriptsize(1.14) & 83.36 \scriptsize (0.43) & 84.15 \scriptsize(1.10) & 84.22 \scriptsize(0.76) & \textbf{84.96 \scriptsize (0.28)} \\ 
    \apparel{} \textrightarrow{} \cameraphoto{} & 91.62 \scriptsize(0.37) & 57.51 \scriptsize(13.32) & 88.56 \scriptsize(0.81) & 87.31 \scriptsize(0.39) & 88.75 \scriptsize (0.69) & \textbf{89.67 \scriptsize(0.32)}  & 88.76 \scriptsize(1.32) & 89.39 \scriptsize (0.23) \\
    \apparel{} \textrightarrow{} \moviereviews{} & 82.08 \scriptsize(0.78) & 35.23 \scriptsize(1.99) & 78.08 \scriptsize(0.46) & 75.54 \scriptsize(0.63) & 76.60 \scriptsize (1.06) & 76.63 \scriptsize(0.92) & 77.39 \scriptsize(0.13) & \textbf{77.63 \scriptsize (0.71} ) \\
    \baby{} \textrightarrow{} \apparel{} & 89.12 \scriptsize(0.38) & 77.52 \scriptsize(11.25) & 87.46 \scriptsize(1.83) & 87.72 \scriptsize(1.85) & 88.47 \scriptsize (0.72) & 88.33 \scriptsize(1.10) & 89.55 \scriptsize(0.10) & \textbf{89.70 \scriptsize (0.23)} \\
    \baby{} \textrightarrow{} \books{} & 86.67 \scriptsize(1.06)  & 43.45 \scriptsize(8.96) & 82.19 \scriptsize(3.70) & 82.89 \scriptsize(3.08) & 83.86 \scriptsize (0.41) & 84.61 \scriptsize(0.39) & 84.38 \scriptsize(0.61) & \textbf{85.01 \scriptsize (0.60} \\ 
    \baby{} \textrightarrow{} \cameraphoto{} & 91.62 \scriptsize(0.37) & 47.58 \scriptsize(7.65) & 89.68 \scriptsize(0.71) & 86.63 \scriptsize(0.53) & 88.73 \scriptsize (0.42) & \textbf{90.63 \scriptsize(0.33)}  & 87.46 \scriptsize(0.88) & 88.64 \scriptsize (0.30) \\
    \baby{} \textrightarrow{} \moviereviews{} & 82.08 \scriptsize(0.78) & 50.63 \scriptsize(7.43) & 77.88 \scriptsize(0.38) & 74.48 \scriptsize(1.79) & 78.07 \scriptsize (0.34) & 78.74 \scriptsize(0.35) & \textbf{79.42 \scriptsize(0.44)} & 78.44 \scriptsize (0.70)\\
    \books{} \textrightarrow{} \apparel{} & 89.12 \scriptsize(0.38) & 37.40 \scriptsize(1.90) & 88.20 \scriptsize(0.51) & 85.90 \scriptsize(0.12) & 85.91 \scriptsize (0.25) & 85.03 \scriptsize(0.36)  & 84.79 \scriptsize(0.75) & \textbf{87.46 \scriptsize (0.27)} \\
    \books{} \textrightarrow{} \baby{} & 87.52 \scriptsize(1.96) & 54.33 \scriptsize(12.49) & 88.56 \scriptsize(0.44) & 82.06 \scriptsize(1.15) & 84.27 \scriptsize (0.11) & 86.50 \scriptsize(0.39)  & \textbf{86.84 \scriptsize(0.48)} & 86.41  \scriptsize (0.79)\\
    \books{} \textrightarrow{} \cameraphoto{} & 91.62 \scriptsize(0.37) & 39.43 \scriptsize(0.49) & 88.58 \scriptsize(1.01) & 86.94 \scriptsize(0.83) & 87.40 \scriptsize (0.44) & 88.44 \scriptsize(0.53) & 87.86 \scriptsize(0.61) & \textbf{88.53 \scriptsize (0.43)} \\
    \books{} \textrightarrow{} \moviereviews{} & 82.08 \scriptsize(0.78) & 54.23 \scriptsize(13.94) & 79.07 \scriptsize(1.01) & 76.19 \scriptsize(0.89) & 79.44 \scriptsize (0.86) & 79.44 \scriptsize(0.95)  & \textbf{80.52 \scriptsize(0.61)}  & 78.91 \scriptsize (0.38)\\
    \cameraphoto{}{} \textrightarrow{} \apparel{} & 89.12 \scriptsize(0.38) & 60.93 \scriptsize(3.78) & 89.76 \scriptsize(0.76) & 87.02 \scriptsize(1.86)& 86.63 \scriptsize (0.29) & 87.74 \scriptsize(1.18)  & 88.53 \scriptsize{(0.42)} & \textbf{88.92 \scriptsize (0.44)} \\
    \cameraphoto{} \textrightarrow{} \baby{} & 87.52 \scriptsize(1.96) & 77.29 \scriptsize(3.61) & 89.42 \scriptsize(0.70) & 88.10 \scriptsize(1.13) & 89.14 \scriptsize (0.30) & 81.71 \scriptsize(2.72)  & \textbf{89.72 \scriptsize(0.43)} & 89.32 \scriptsize (0.42) \\ 
    \cameraphoto{} \textrightarrow{} \books{} & 86.67 \scriptsize(1.06) & 38.21 \scriptsize(1.40) & 85.56 \scriptsize(0.62) & 81.18 \scriptsize(2.07) & 83.61 \scriptsize (0.67) & 80.55 \scriptsize(0.81) & 84.14 \scriptsize(0.52) & \textbf{85.42 \scriptsize (0.70)} \\
    \cameraphoto{} \textrightarrow{} \moviereviews{} &  82.08 \scriptsize(0.78) & 35.08 \scriptsize(1.94) & 76.13 \scriptsize(0.54) & 64.99 \scriptsize(5.91) & \textbf{74.22 \scriptsize (0.31)} & 69.53 \scriptsize(1.24)  & 73.22 \scriptsize(0.48) & 73.50 \scriptsize (0.84) \\
    \moviereviews{} \textrightarrow{} \apparel{} & 89.12 \scriptsize(0.38) & 37.07 \scriptsize(4.16) & 82.64 \scriptsize(2.17) & 81.05 \scriptsize(1.15) & 79.56 \scriptsize (0.53) & 82.45 \scriptsize(1.43) & 81.93 \scriptsize(0.47) & \textbf{84.41 \scriptsize (0.43)} \\
    \moviereviews{} \textrightarrow{} \baby{} & 87.52 \scriptsize(1.96) & 38.76 \scriptsize(4.17) & 80.59 \scriptsize(2.18) & 77.95 \scriptsize(1.46) & 79.33 \scriptsize (0.43) & 81.70 \scriptsize(1.22) & 84.28 \scriptsize(0.41) & \textbf{84.91 \scriptsize (0.36)} \\ 
    \moviereviews{} \textrightarrow{} \books{} & 86.67 \scriptsize(1.06) & 42.07 \scriptsize(4.86) & 85.13 \scriptsize(0.83) & 82.83 \scriptsize(0.62) & \textbf{84.90 \scriptsize(1.29)} & \textbf{84.90 \scriptsize (0.23)}  & 84.47 \scriptsize(0.80) & 84.45 \scriptsize (0.31) \\ 
    \moviereviews{} \textrightarrow{} \cameraphoto{} & 91.62 \scriptsize(0.37) & 36.92 \scriptsize(1.86) & 86.56 \scriptsize(0.63) & 84.58 \scriptsize(0.46) & 82.53 \scriptsize (0.92) & 86.68 \scriptsize(0.65)  & 86.25 \scriptsize(0.38) & \textbf{88.37 \scriptsize (0.11)} \\  [0.5ex] \hline
    Avg & 87.40 \scriptsize(0.91)  & 49.28 \scriptsize(5.47) & 84.92 \scriptsize(1.03) & 81.91 \scriptsize (1.37)  & 83.68 \scriptsize(0.50) & 83.72 \scriptsize(0.88)  & 84.60 \scriptsize(0.57) & \textbf{85.16 \scriptsize (0.43)} \\
 \hline
 \end{tabular} %
 }
 \caption{\label{tab:amazon_results}F1 scores for \amazondataset{} dataset. We report mean and standard deviation of 3 runs. The five domains are Apparel (\apparel{}), Baby (\baby{}), Books (\books{}), Camera\_Photo (\cameraphoto{}) and Movie Reviews (\moviereviews{}). On average, our method outperforms all baselines. Our methods are competitive with fully unsupervised domain adaptation methods.}
\end{table*}

\subsection{Datasets}
We evaluate our approach on two representative datasets with different tasks, both in English.  \Cref{tab:dataset} shows the details of the datasets. Every dataset has 5 domains, and we consider each domain with every other domain which results in 20 domain adaptation scenarios per dataset, 120 experiments per method, totalling over 1.9K experiments.

\noindent
\paragraph{\amazondataset:} Multi Domain Sentiment Analysis Dataset \cite{blitzer-etal-2007-biographies} that contains Amazon product reviews for five different types of products (domains): Apparel (\apparel{}), Baby (\baby{}), Books (\books), Camera\_Photo (\cameraphoto{}), and Movie Reviews (\moviereviews{}). Each review is labeled as positive or negative. We follow the setup in \citep{ramesh-kashyap-etal-2021-domain}.

\noindent
\paragraph{\mnlidataset{}:} The Multigenre Natural Language Inference (MNLI) corpus \cite{williams-etal-2018-broad}
contains hypothesis--premise pairs covering a variety of genres: Travel (\travel{}), fiction (\fiction{}), telephone (\telephone{}), government (\government{}), and slate (\slate{}). Each pair of sentences is labeled Entailment, Neutral, or Contradiction. The train and validation data set are taken from the train set by sampling 90\% and 10\% samples, respectively. We use the MNLI-matched validation set as our test set.

\subsection{Baseline Methods} 

\noindent
\paragraph{Fully supervised.} 
\textit{Fine-tune (\fireemoji)}:  Fine-tunes a language model using labeled data from the target domain. Serves as an upper bound of performance.

\noindent
\paragraph{Unsupervised Domain Adaptation (\uda{}).}
\textit{Domain Adversarial Neural Networks} (\dann{}): An unsupervised domain adaptation method \cite{dann} that learns domain-invariant information by minimizing task loss and maximizing domain confusion loss with the help of gradient reversal layers. \textit{Domain Separation Networks:} (\dsn{}) \cite{dsn} improves \dann{}, with additional losses to preserve domain-specific information along with the extraction of domain-invariant information. \bertbase{} serves as a feature extractor for both methods.

\noindent
\paragraph{Adapter Based.}
\textit{\dann{} Adapter} (\dannadapter{}): Similar to \dann{}, but we insert trainable adapter modules into every layer of a PLM. \textit{\dann{} Adapter with Multiple Classifiers} (\dannadaptermc{}): Unlike \dann{}-\adapteremoji which involves a single task and domain classifier,  here a task and domain classifier are added to each of the last 3 layers of a PLM. The representation of the last layers of a PLM is domain variant \cite{ramesh-kashyap-etal-2021-analyzing}, and this model obtains domain-invariant information\footnote{We tried adding classifiers incrementally to the last few layers. Adding it to the last 3 layers performed the best.} 
(vi) Task adapter (\taskadapter{}): Adapter fine-tuning \cite{pfeiffer-etal-2020-adapterhub} where adapters are fine-tuned in the labeled source domain and tested in the target domain. (vii) Two-step Domain and Task Adapter (\ourmethod{}): This work, where we first train a domain adapter that reduces the probabilistic divergence between two domains and then fine-tunes a task adapter by stacking. (viii) Joint Domain Task Adapter (\jointdt) - We train a single adapter that reduces the domain and task loss jointly. For all adapter-based experiments, the PLM is frozen, and only adapter modules are trained. 

Since we use adapters, we only consider other adapter based baselines and omit other methods such as Prefix-tuning \cite{lester-etal-2021-power}. Also, \citep{Zhang2021UnsupervisedDA} target multidomain adaptation and use data from all the domains during training unlike our method and is not a fair comparison.

\paragraph{Implementation Details and Evaluation.}
For our experiments, we use \bertbase{} \cite{devlin-etal-2019-bert} available in the HuggingFace Transformers library \cite{wolf-etal-2020-transformers} as our backbone.  Adapter implementations are from AdapterHub \cite{pfeiffer-etal-2020-adapterhub}. We follow \citep{pfeiffer-etal-2021-adapterfusion} and add only one bottleneck layer after the feedforward layer.

We use the AdamW optimizer and a learning rate of $1e-4$ for all our adapter-based training and $2e-5$ otherwise. Only for the smaller \amazondataset{} dataset, we used an adapter bottleneck size (reduction factor) of 32. For all other adapter-based experiments and datasets, we use the default adapter bottleneck size of 16. We performed experiments on three different seeds. We report the mean and standard deviation of the F1 scores.  For \dann{} we use 0.04 as our 
$\lambda$ and for \dsn{} we use 0.1, 0.1, and 0.3 as our weights for three losses: reconstruction, similarity, and difference respectively. We avoid extensive hyperparameter tuning per domain adaptation scenario for efficiency.

\renewcommand{\arraystretch}{1.2}
\begin{table*}[t!]
\centering \footnotesize 
\resizebox{\linewidth}{!}{%
 \begin{tabular}{l|c|@{\hskip 0.4in}cc|ccccc} 
 \hline
 & \colorbox{red!30}{\textbf{Fully Supervised}} & \multicolumn{2}{@{\hskip -0.3in}c@{\hskip 0.1in}|}{\colorbox{orange}{\textbf{Unsupervised Domain Adaptation}}} & \multicolumn{5}{c}{\colorbox{blue!25}{\textbf{Adapter Based}}} \\ 
 Src \textrightarrow Trg & \fireemoji & \dann{} & \dsn{} & \dannadapter{} & \dannadaptermc{} & \taskadapter{} & \ourmethod{} & \jointdt{} \\ [0.5ex] 
 \hline
    \fiction{} \textrightarrow \slate{} & 74.09 \scriptsize(0.40) & 73.68 \scriptsize(0.21) & 72.36 \scriptsize(0.17) & 70.96 \scriptsize(0.03) & 62.40 \scriptsize(4.79) & 72.36 \scriptsize(0.36) & \textbf{73.46 \scriptsize(0.34)} & 72.30 \scriptsize(0.26)  \\
    \fiction{} \textrightarrow \government{} & 82.19 \scriptsize(0.12) & 79.17 \scriptsize(0.25) & 79.79 \scriptsize(0.21) & 78.73 \scriptsize(0.43) & 77.23 \scriptsize(0.33) & 79.00 \scriptsize(0.46) & 78.65 \scriptsize(0.25) & \textbf{79.79 \scriptsize(0.22)} \\
    \fiction{} \textrightarrow \telephone{} & 78.41 \scriptsize(0.66) & 73.72 \scriptsize(0.81) & 75.07 \scriptsize(0.32) & 70.89 \scriptsize(0.74) & 71.68 \scriptsize(0.59) & 70.83 \scriptsize(0.54) & \textbf{73.05 \scriptsize(0.70)} & 71.59 \scriptsize(0.78) \\
    \fiction \textrightarrow \travel{} & 81.81 \scriptsize(0.20) & 76.99 \scriptsize(0.19) & 76.82 \scriptsize(0.50) & 74.42 \scriptsize(0.18) & 75.09 \scriptsize(0.05) & 75.85 \scriptsize(0.19) & 76.75 \scriptsize(0.80) & \textbf{77.07 \scriptsize(0.26)} \\
    \slate{} \textrightarrow \fiction{} & 78.59 \scriptsize(0.34) & 75.91 \scriptsize(0.23) & 76.62 \scriptsize(0.38) & 73.89 \scriptsize(0.61) & 73.47 \scriptsize(0.28) & 75.25 \scriptsize(0.19) & \textbf{75.52 \scriptsize(0.89)} & 75.35 \scriptsize(0.56) \\
    \slate{} \textrightarrow \government{} & 82.19 \scriptsize(0.12) & 80.91 \scriptsize(0.46) & 81.27 \scriptsize(0.23) & 79.99 \scriptsize(0.36) & 79.16 \scriptsize(0.10) & 80.76 \scriptsize(0.40) & \textbf{81.65 \scriptsize(0.11)} & 80.94 \scriptsize(0.30) \\
    \slate{} \textrightarrow \telephone{} & 78.41 \scriptsize(0.66) & 74.32 \scriptsize(0.57) & 74.27 \scriptsize(0.48) & 72.29 \scriptsize(0.57) & 71.89 \scriptsize(0.07) & 72.66 \scriptsize(0.79) & \textbf{74.09 \scriptsize(0.30)} & 73.38 \scriptsize(0.63) \\
    \slate{} \textrightarrow \travel{} & 81.81 \scriptsize(0.20) & 76.81 \scriptsize(0.35) & 78.17 \scriptsize(0.20) & 75.58 \scriptsize(0.54) & 75.77 \scriptsize(0.39) & 76.16 \scriptsize(0.22) & \textbf{77.31 \scriptsize(0.60)} & 77.16 \scriptsize(0.18) \\
    \government{} \textrightarrow \fiction{} & 78.59 \scriptsize(0.34) & 73.41 \scriptsize(0.73) & 72.62 \scriptsize(0.37) & 71.57 \scriptsize(0.68) & 70.34 \scriptsize(0.73) & 72.66 \scriptsize(0.31) & 72.66 \scriptsize(0.56) & \textbf{73.56 \scriptsize(0.23)} \\
    \government{} \textrightarrow \slate{} & 74.09 \scriptsize(0.40) & 72.51 \scriptsize(0.10) & 71.93 \scriptsize(0.25) & 70.17 \scriptsize(0.64) & 69.49 \scriptsize(0.40) & 71.11 \scriptsize(0.38) & 71.14 \scriptsize(0.21) & \textbf{71.36 \scriptsize(0.04)} \\
    \government{} \textrightarrow \telephone{} & 78.41 \scriptsize(0.66) & 71.52 \scriptsize(0.13) & 72.90 \scriptsize(0.39) & 69.45 \scriptsize(0.96) & 68.67 \scriptsize(0.17) & 71.40 \scriptsize(0.30) & 71.53 \scriptsize(1.04) & \textbf{71.99 \scriptsize(0.67)} \\
    \government{} \textrightarrow \travel{} & 81.81 \scriptsize(0.20) & 77.42 \scriptsize(0.54) & 77.80 \scriptsize(0.42) & 74.35 \scriptsize(0.22) & 74.04 \scriptsize(0.51) & 76.29 \scriptsize(0.10) & 76.16 \scriptsize(0.34) & \textbf{76.79 \scriptsize(0.59)} \\
    \telephone{} \textrightarrow \fiction{} & 78.59 \scriptsize(0.34) & 75.07 \scriptsize(0.08) & 75.17 \scriptsize(0.35) & 72.24 \scriptsize(0.59) & 71.49 \scriptsize(0.45) & \textbf{74.48 \scriptsize(0.33)} & 73.34 \scriptsize(0.41) & 73.89 \scriptsize(0.12) \\
    \telephone{} \textrightarrow \slate{} & 74.09 \scriptsize(0.40) & 71.65 \scriptsize(0.50) & 72.16 \scriptsize(0.23) & 69.09 \scriptsize(1.79) & 69.25 \scriptsize(0.31) & 70.94 \scriptsize(0.16) & 70.94 \scriptsize(0.55) & \textbf{71.41 \scriptsize(0.19)} \\
    \telephone{} \textrightarrow \government{} & 82.19 \scriptsize(0.12) & 78.57 \scriptsize(0.60) & 79.24 \scriptsize(0.31) & 77.80 \scriptsize(0.27) & 76.65 \scriptsize(0.20) & 79.24 \scriptsize(0.35) & 79.65 \scriptsize(0.60) & \textbf{79.78 \scriptsize(0.64)} \\
    \telephone{} \textrightarrow \travel{} & 81.81 \scriptsize(0.20) & 75.72 \scriptsize(0.37) & 77.29 \scriptsize(0.61) & 74.67 \scriptsize(0.50) & 74.08 \scriptsize(0.25) & 75.27 \scriptsize(0.83) & \textbf{76.11 \scriptsize(0.91)} & 75.95 \scriptsize(0.50) \\
    \travel{} \textrightarrow \fiction{} & 78.59 \scriptsize(0.34) & 73.22 \scriptsize(0.92) & 72.44 \scriptsize(0.50) & 70.27 \scriptsize(0.45) & 69.08 \scriptsize(0.64) & 72.20 \scriptsize(0.49) & 73.12 \scriptsize(0.08) & \textbf{73.13 \scriptsize(0.22)} \\
    \travel{} \textrightarrow \slate{} & 74.09 \scriptsize(0.40) & 70.76 \scriptsize(0.72) & 70.97 \scriptsize(0.26) & 68.35 \scriptsize(0.62) & 67.23 \scriptsize(0.39) & 70.28 \scriptsize(0.37) & 70.67 \scriptsize(0.50) & \textbf{71.28 \scriptsize(0.38)} \\
    \travel{} \textrightarrow \government{} & 82.19 \scriptsize(0.12) & 80.91 \scriptsize(0.28) & 81.67 \scriptsize(0.37) & 79.25 \scriptsize(0.34) & 78.77 \scriptsize(0.32) & 81.26 \scriptsize(0.37) & 81.11 \scriptsize(0.42) & \textbf{81.55 \scriptsize(0.16)} \\
    \travel{} \textrightarrow \telephone{} & 78.41 \scriptsize(0.66) & 70.41 \scriptsize(1.63) & 71.98 \scriptsize(0.50) & 69.33 \scriptsize(0.41) & 69.45 \scriptsize(0.39) & 70.98 \scriptsize(0.11) & 70.95 \scriptsize(0.19) & \textbf{71.42 \scriptsize(0.12)} \\[0.5ex] \hline
    Avg & 79.02 \scriptsize(0.34) & 75.13 \scriptsize(0.48) & 75.53 \scriptsize(0.35) & 73.16 \scriptsize(0.55) & 72.26 \scriptsize(0.57) & 74.45 \scriptsize(0.40) & \textbf{74.89 \scriptsize(0.49)} & \textbf{74.98 \scriptsize(0.35)} \\
 \hline
 \end{tabular} %
 }
 \caption{\label{tab:mnli_results}F1 scores for \mnlidataset{} dataset. We report mean and standard deviation of 3 runs. The five domains are Fiction (\fiction{}), Slate (\slate{}), Government (\government{}), Telephone (\telephone{}), and Travel (\travel{}). On average, our method performs better than all baselines. }
\end{table*}

\subsection{Results}
From \Cref{tab:amazon_results} and \Cref{tab:mnli_results} our methods \ourmethod{} and \jointdt{} perform well in both \amazondataset{} and \mnlidataset{}. We find that fine-tuning the task adapter (\taskadapter{}) is a strong baseline and, compared to it, we perform well in 17/20 domain adaptation scenarios in \amazondataset{} (largest increase of 8 points for \cameraphoto{} \textrightarrow{} \baby{} ) and 19/20 domain adaptation scenarios in \mnlidataset{} (largest increase of 2.2 for \fiction{} \textrightarrow{} \telephone{}). 
One possible explanation of scenarios where our method finds the largest increase is the proximity of the two domains. The overlap in vocabularies (\Cref{fig:vocab_overlap} in the Appendix) between \cameraphoto{} \textrightarrow \baby{} in \amazondataset{} and \fiction{} \textrightarrow{} \telephone{} in \mnlidataset{} is high, and our method takes advantage of learning domain-invariant information that can be used for efficient domain transfer. Our methods for learning domain-invariant information are necessary to achieve good domain adaptation.

\paragraph{\systemname{} is comparable to \uda{} methods.} Compared to \uda{} methods where all parameters of the backbone model are fine-tuned, we perform close to them on average. \jointdt{} performs better than \dsn{} by 0.2\% in \amazondataset{}. We are within 0.85\% in \mnlidataset{} compared to \dsn{}. Training \dann{} is highly unstable and produces varied results, especially for \amazondataset{} with a small number of examples in each domain. Our adapter method achieves better results compared to \dann{} with a minimal modification of the hyperparameters.

\paragraph{Replacing \uda{} Feature Extractors with Adapter Versions is insufficient.}

\textit{Given that fully fine-tuned \uda{} methods perform well, can we freeze the feature extractors \uda{} methods and \ft{} only adapters and perform effective domain adaptation?} We compare our methods with \dannadapter{} and \dannadaptermc{} and outperform them both in \amazondataset{} and \mnlidataset{}. This is in line with \citet{karouzos-etal-2021-udalm} that although domain adversarial training  brings domain representations closer, it introduces distortion in the semantic space, reducing model performance. This shows that simply replacing feature extractors with their adapter versions in existing \uda{} methods is not an effective strategy.

\paragraph{Gap to Full Fine-Tuning.}
Fine-tuning a PLM with supervised data in the target domain is the upper bound performance for domain adaptation. The gap from full \fting{} is greater when more data are available (3.15 in \amazondataset{} and 4.13 in \mnlidataset{}). This is not surprising, as the supervised \fting{} works better with more data. However, while adapters perform closely to complete fine-tuning in supervised scenarios \cite{towards-a-unified-view-of-parameter-efficient-transfer-learning}, there is still a large gap between domain adaptation and complete fine-tuning.

\subsection{Further Analysis}
\label{sec:further-analysis}

\begin{figure*}[t!]
    \centering
    \subfloat[ \label{fig:rf_ablation_amazon}]{\includegraphics[width=0.53\textwidth]{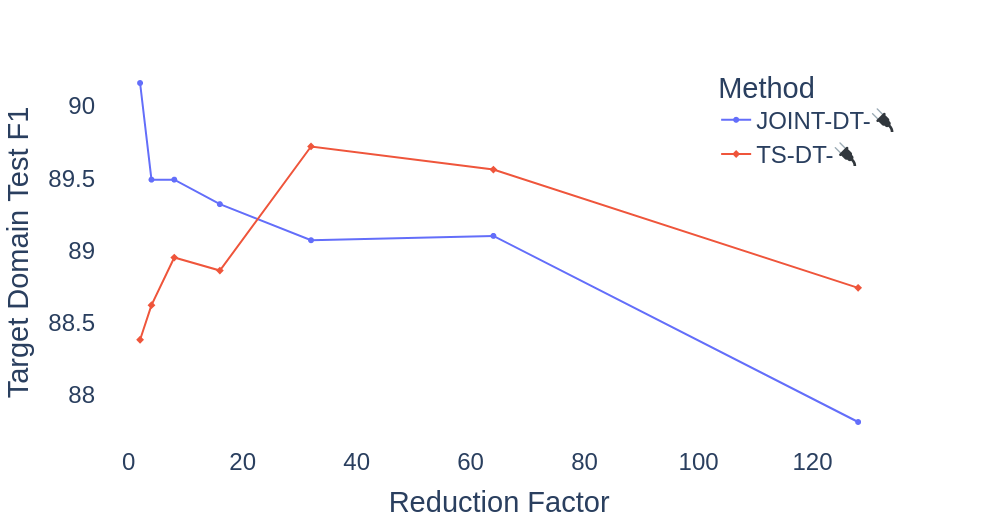}} 
    \subfloat[\label{fig:rf_ablation_mnli}]{\includegraphics[width=0.53\textwidth]{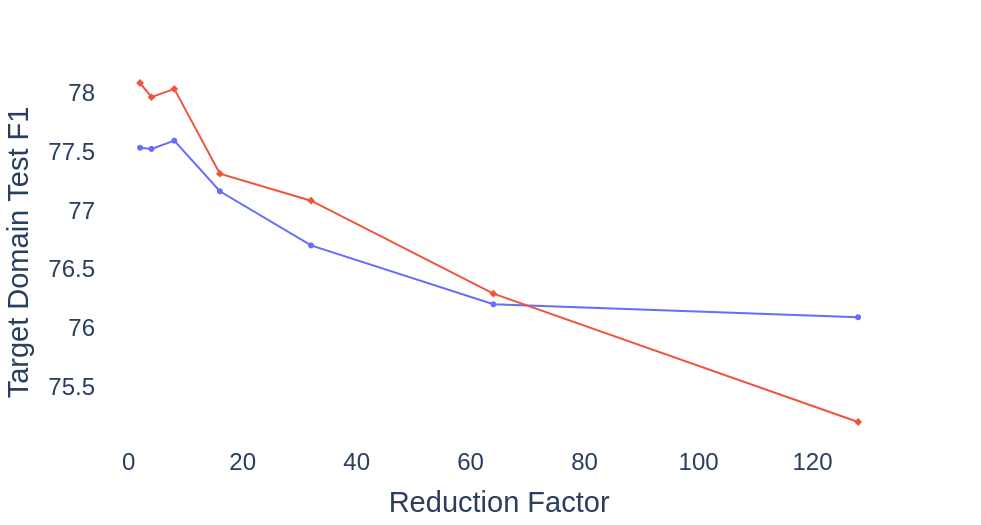}} 
    \caption{(a) Performance for \amazondataset{} on the \cameraphoto{} \textrightarrow{} \baby{} domain adaptation scenario for different reduction factors. (b) Performance for \mnlidataset{} on the \slate{} \textrightarrow{} \travel{} scenario for different reduction factors.}
    \label{fig:ablation_rf}
\end{figure*}

\paragraph{Adapter Reduction Factor.} The bottleneck size ($d$) of the adapters plays an important role in the final performance of the model. We show the performance of the models at various reduction factors in \Cref{fig:ablation_rf}. For \jointdt{}, smaller reduction factors generally perform well in both \amazondataset{} and \mnlidataset{}, with performance reducing for larger reduction factors. This shows that the \jointdt{} method requires a greater number of parameters to reduce divergence and learn task representations together. Since \ourmethod{} adds two adapters, this increases the number of parameters added for the same reduction factor compared to \jointdt{}. As a result, we find that as the data scale up, relatively low reduction factors work well.

\begin{figure}[ht!]
  \begin{tabular}{l}
   \includegraphics[trim = 5mm 0mm 15mm 0mm, clip, width=.5\textwidth, right]{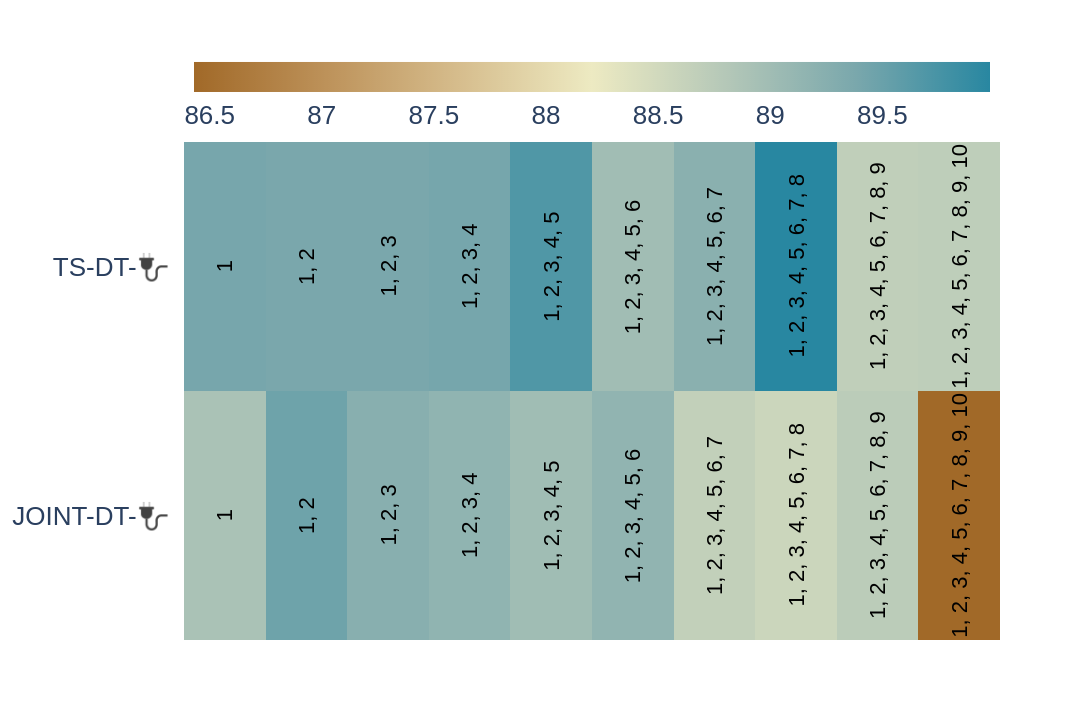} \vspace{-0.5cm}\\    
   \includegraphics[trim = 5mm 0mm 15mm 0mm, clip, width=.5\textwidth, right]{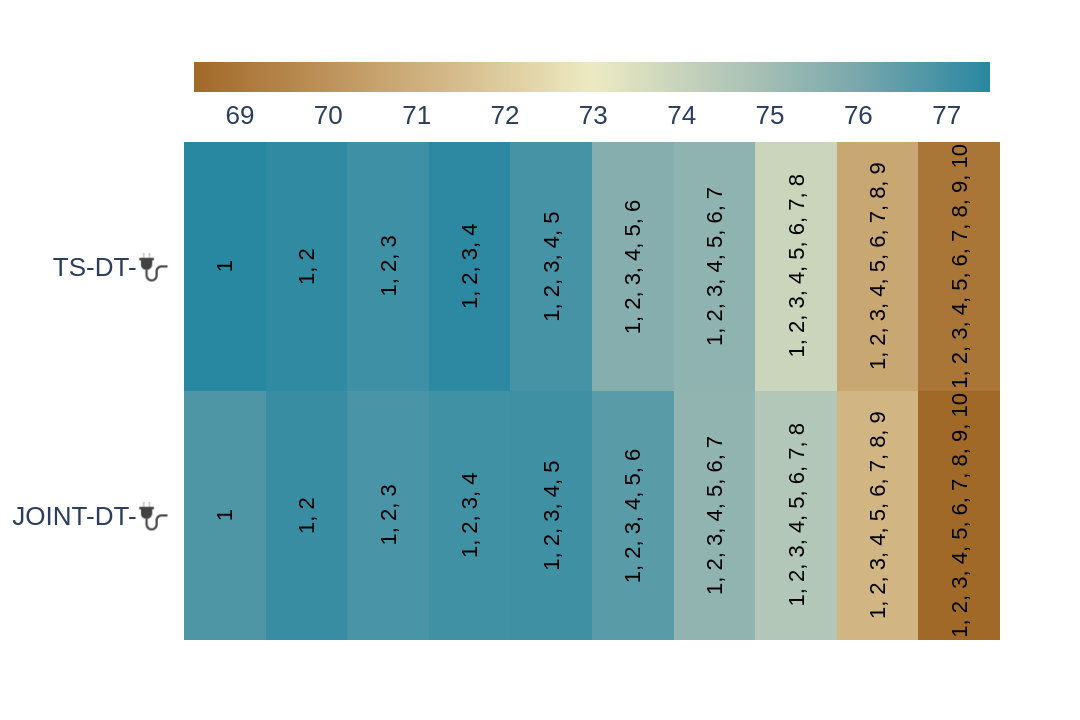} \\  
  \end{tabular}
  \caption{Difference in performance when adapters are removed from certain layers (mentioned inside the cells) for the \amazondataset{} dataset (top) and for \mnlidataset{} dataset (bottom). The performance reduces if adapters are removed from certain layers }
  \label{fig:skip_layer_rf}
\end{figure}

\paragraph{The removal of adapters from continuous layer spans.} All adapters are not equal. Removing adapters from the first few layers still preserves performance (\Cref{fig:skip_layer_rf}).
For \jointdt{} and \ourmethod{}, the F1 slowly decreases as we continually remove the adapters. However, we obtained a comparable performance after removing the adapters from layers 1-6. This suggests that adapters are effective when added to higher layers, where the divergence between domains is greater at higher layers compared to lower layers \cite{ramesh-kashyap-etal-2021-analyzing}. Thus we can further reduce the number of parameters for domain adaptation.

\begin{figure*}[t!]
\centering \footnotesize 
\resizebox{\linewidth}{!}{%
 \begin{tabular}{c@{}c@{}c@{}c@{}c@{}}
 \includegraphics[width=\textwidth]{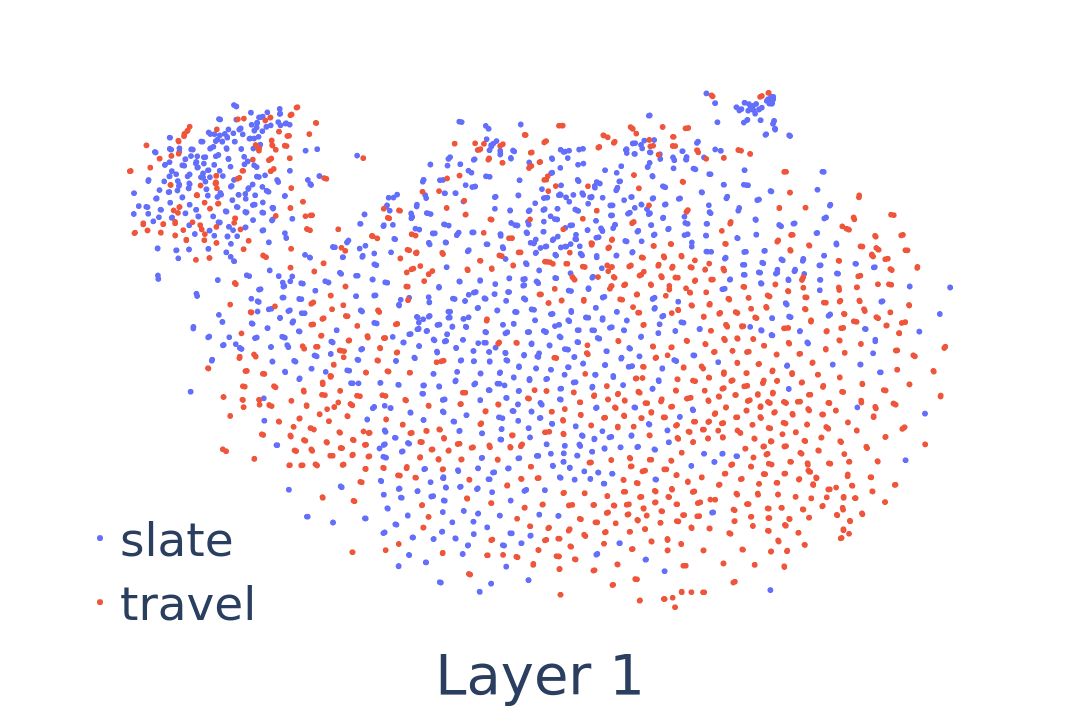}&
     \includegraphics[width=\textwidth]{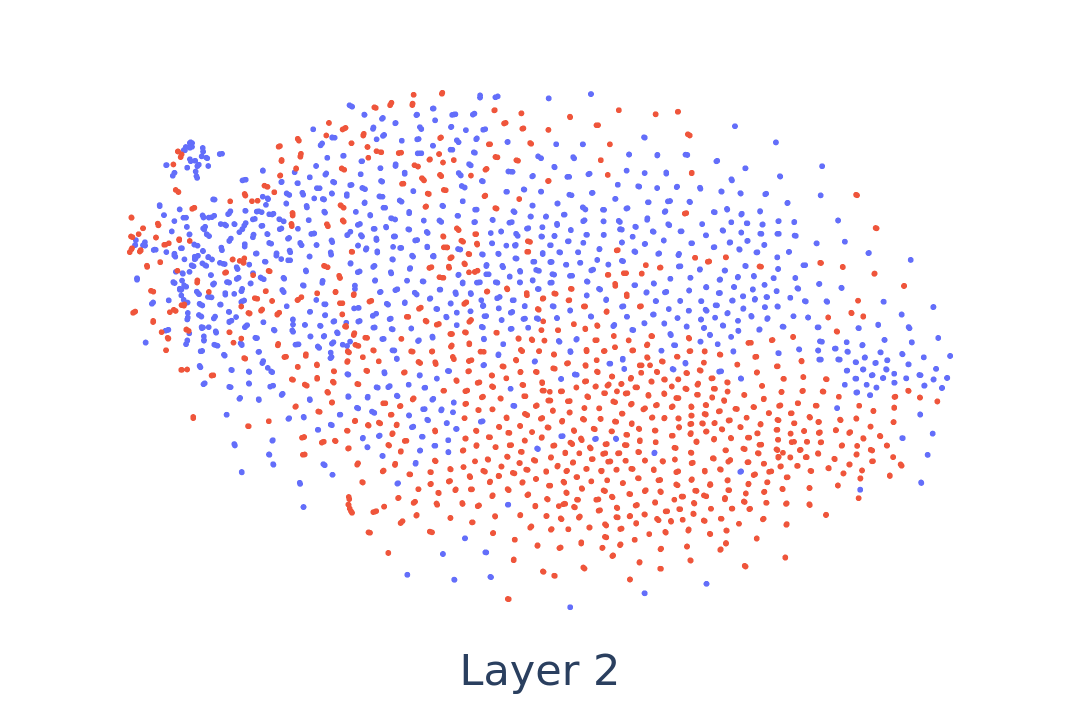}&
     \includegraphics[width=\textwidth]{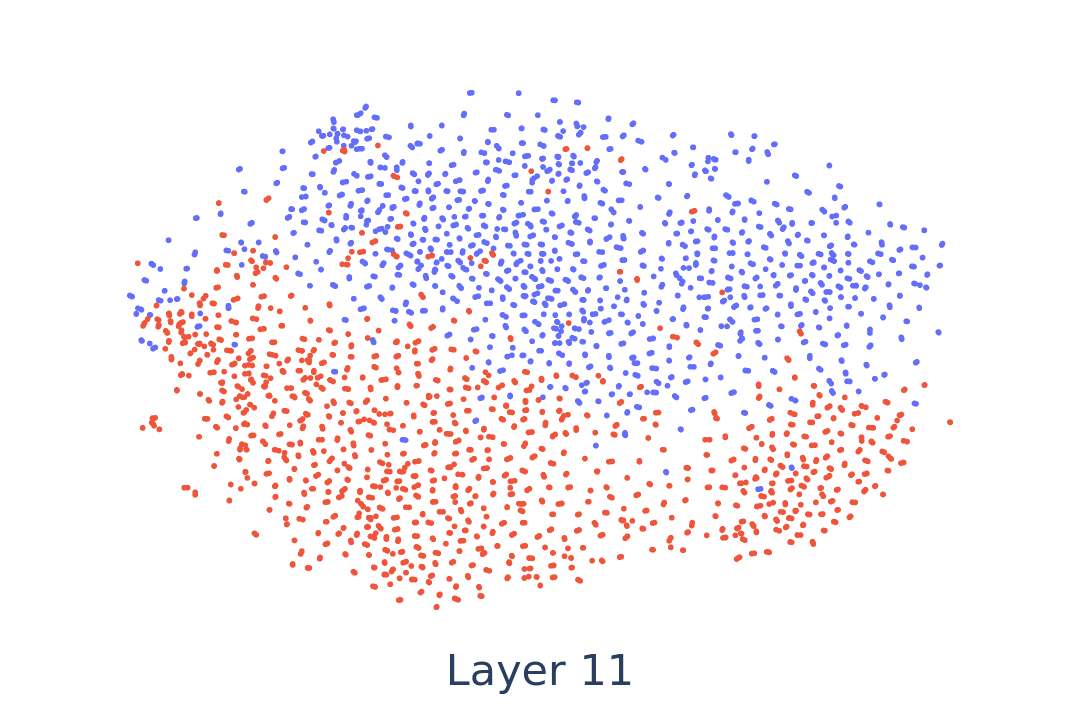}&
     \includegraphics[width=\textwidth]{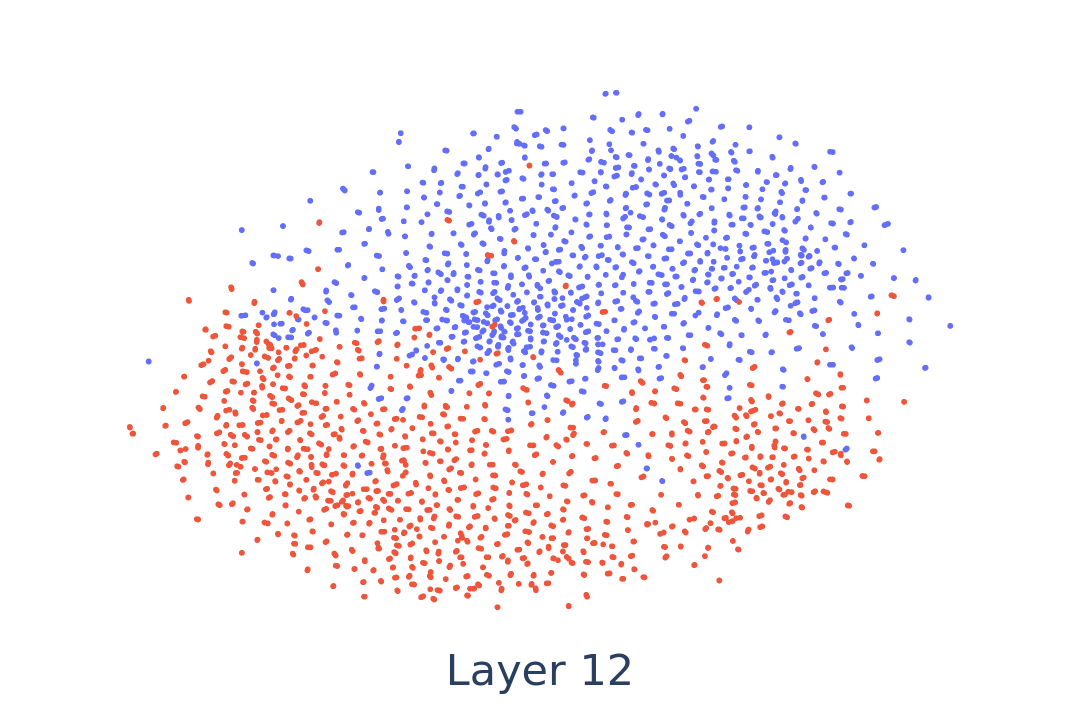}  \\ 
     \includegraphics[width=\textwidth]{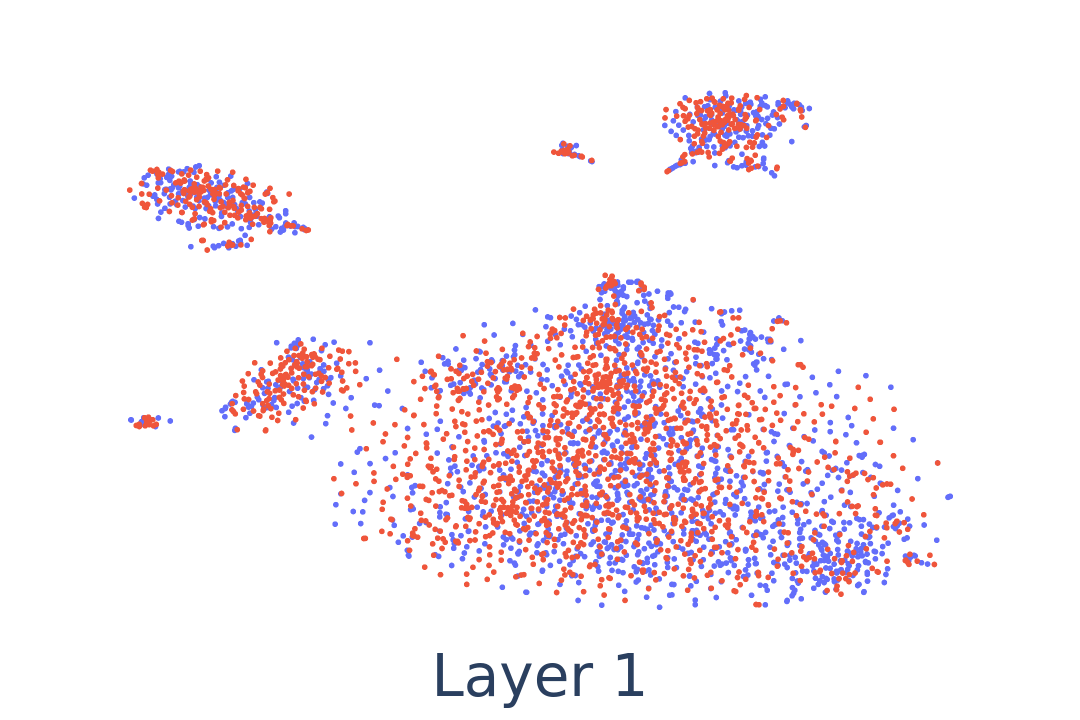}&
     \includegraphics[width=\textwidth]{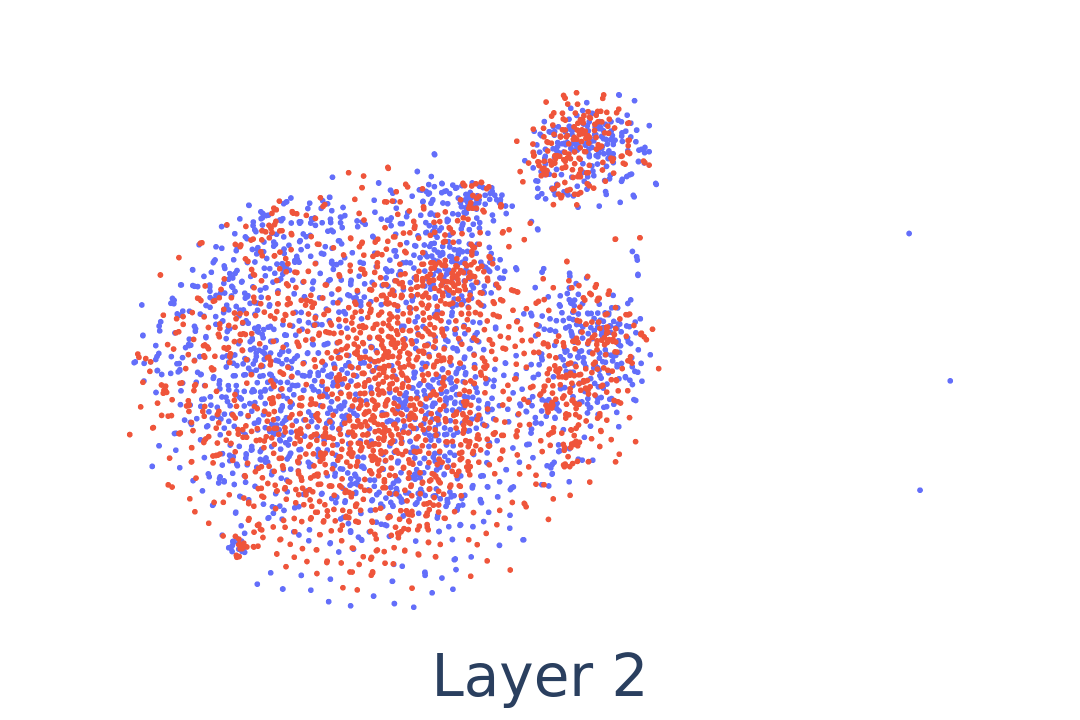}&
     \includegraphics[width=\textwidth]{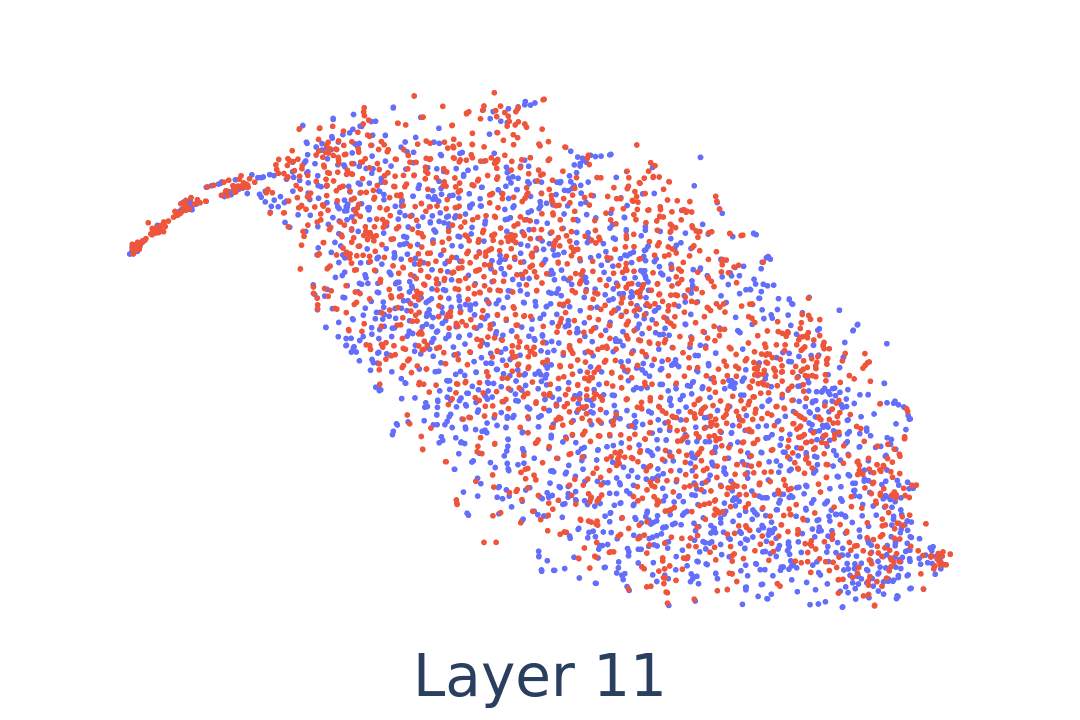}&
     \includegraphics[width=\textwidth]{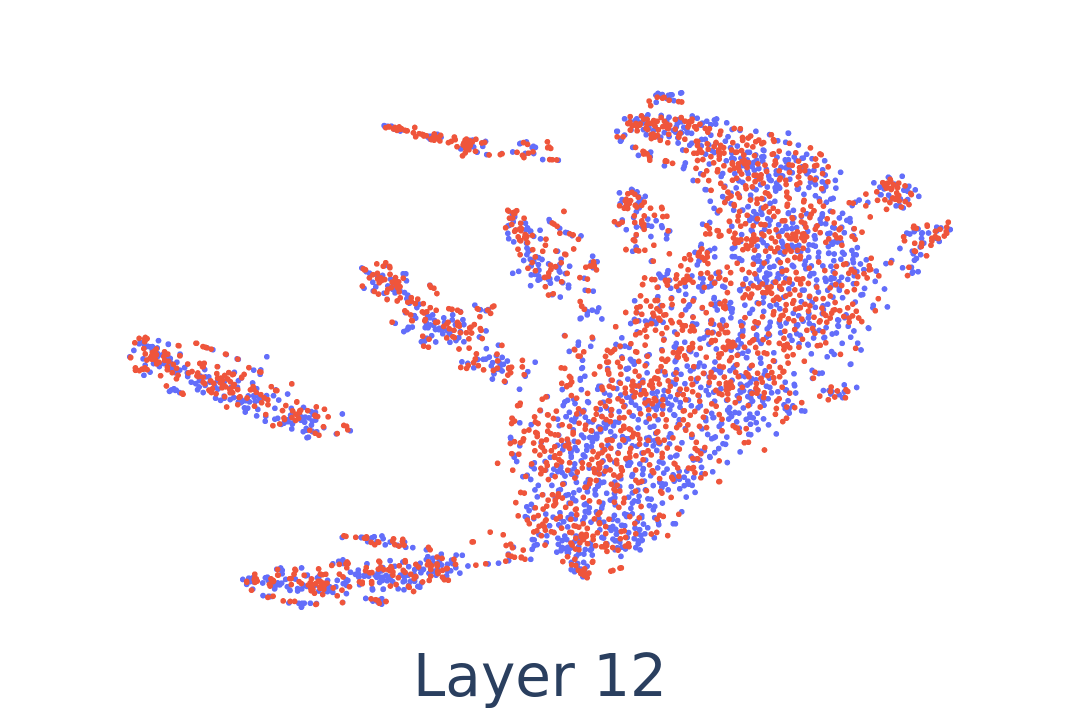}\\

  \end{tabular} %
  }
 \caption{(top) t-SNE plots for the representations from \bertbase{}. The lower layers are domain invariant while the higher layers are domain-variant (bottom) tSNE plots from the domain adapter trained on the \slate{} \textrightarrow{} \travel{} domain. We reduce the divergence using domain adapters where even higher layers are domain invariant. }
 \label{fig:domain-adapter-tsne}
\end{figure*}

\paragraph{t-SNE plots.}
The t-SNE \cite{JMLR:v9:vandermaaten08a} plots from domain adapters are shown in \Cref{fig:domain-adapter-tsne} for the data set \mnlidataset{}. The lower layers have low divergence and the data from the two domains are interspersed, whereas the higher layers have high divergence. Our method effectively reduces the divergence in higher layers.

\paragraph{Composability.} We test the composability of our two-step method \ourmethod{}. We reuse the task adapter trained for \cameraphoto{} \textrightarrow{} \baby{} and replace the domain adapter with the domain adapter of \cameraphoto{} \textrightarrow{} \moviereviews{} and perform inference on \cameraphoto{} \textrightarrow{}
\moviereviews{} dataset. The initial F1 of the \cameraphoto{} \textrightarrow{} \moviereviews{} dataset was 73.22 and after composing it with a different task adapter, the F1 score is 72.66 -- a minimal performance loss. This shows the composability of \ourmethod{}. 


\section{Literature Review}

\paragraph{Parameter Efficient Fine-tuning Methods.}

Adapters \cite{pmlr-v97-houlsby19a} are task-specific modules added to frozen transformer layers, with only the adapter parameters updated. Their plug-and-play characteristics and the avoidance of catastrophic forgetting have resulted in their use for NLP tasks: machine translation \cite{bapna-firat-2019-simple}, named entity recognition \cite{pfeiffer-etal-2020-mad}, etc. Recently, \cite{he-etal-2021-effectiveness} have shown that they are efficient in scenarios where there is minimal supervised data. However, they neither test their performance under domain shift nor propose methods to improve adapter \fting{}. Closely related to our method is the work of \citet{ngo-trung-etal-2021-unsupervised}, who learns a shared-private representation per layer, similar to \dsn{} \cite{dsn}. Their method requires balancing multiple loss functions, compared to our simpler two-step domain adaptation method. The stacking of adapters has been followed before by \cite{pfeiffer-etal-2020-mad} for cross-lingual tasks: learning a language adapter first and stacking a task adapter. However, one language adapter is learned per language, assumes large amounts of unsupervised data to be available in all the languages, and requires supervised data to be available to learn a task, which is not applicable for domain adaptation. Compared to other methods, we make domain adaptation more efficient using principles of unsupervised domain adaptation.

\paragraph{Unsupervised Domain Adaptation (\uda{}).} Existing \uda{} approaches can be categorized into model-centric, data-centric, and hybrid. \textit{Model-centric} approaches involve augmenting feature space or altering the loss function, architecture, or model parameters \cite{blitzer-etal-2006-domain,10.1145/1772690.1772767, dann} have been popular. 
A popular \textit{model-centric} approach is to use adversarial training between the domain and the task classifier \cite{dann} to extract domain-invariant information. \cite{dsn} in addition preserves domain-specific information. These works involve training a large number of parameters and require careful balancing of multiple loss functions. Our methods build on top of these works and make it more parameter-efficient.

Large-scale transformers pretrained on domain-specific corpora have been a norm: biomedical \cite{10.1093/bioinformatics/btz682}, scientific publications \cite{beltagy-etal-2019-scibert}, among others. Another alternative is to continue pretraining generic models on domain-specific data: domain adaptive pretraining \cite{gururangan-etal-2020-dont}. Both solutions are expensive since a huge model has to be stored for every domain while using adapters affords storing a small number of parameters for every domain pair and can be quickly adapted to new domains.

\section{Discussion}
This work shows that domain adaptation in NLP can be made more efficient using adapters. We use adapters \fting{} \cite{pmlr-v97-houlsby19a} proposed before and stacking of adapters that have been proposed before for a cross-lingual setting \cite{pfeiffer-etal-2020-mad} for the unsupervised domain adaptation. The approach we have discussed will make domain adaptation more practical for real-world use cases, making adaptation faster and cheaper. However, in this work, we have used \bertbase{} for all of our methods. Using other backbone transformer models is part of our future work. We deal only with a classification and natural language inference task. Adapters have previously been used for machine translation \cite{bapna-firat-2019-simple} and other generation tasks \cite{zhang-etal-2022-continual}. We need to explore our domain adaptation methods for other generation tasks. 

In this work, we reduce the marginal distribution of the two distributions. Previous works such as \citet{coregularized-domain-adaptation} show that reducing only the marginal distribution is not sufficient and aligning the label distributions is necessary. However, NLP works do not consider this and would require further investigation by the community.

\section{Conclusion}
In this work, we propose \systemname{}, to make unsupervised domain adaptation more parameter-efficient. Our  methods outperform other strong baselines, and we show that we can perform better than just training a task adapter on supervised data. We perform competitively to other \uda{} methods at a fraction of the parameters and outperform them when there is limited data -- a more practical scenario. Future work should explore other parameter-efficient methods such as prefix-tuning \cite{li-liang-2021-prefix} for domain adaptation. NLP should also consider other avenues, such as continuous adaptation to new domains and adaptation to new domains when there are no data available.

\section{Acknowledgments}
This research is supported by the SRG grant id: T1SRIS19149 and the Ministry of Education, Singapore, under its AcRF Tier-2 grant (Project no. T2MOE2008, and Grantor reference no. MOET2EP20220-0017). Any opinions, findings, conclusions, or recommendations expressed in this material are those of the author(s) and do not reflect the views of the Ministry of Education, Singapore.

\section{Limitations}
We have several limitations to our work. We have experimented with only one type of parameter-efficient method, which is the adapter \fting{} method. Several other alternative parameter-efficient methods, such as LoRA \cite{lora}, Bitfit \cite{bitfit}, and other unifying paradigms \cite{towards-a-unified-view-of-parameter-efficient-transfer-learning}, have been proposed in recent times. These methods are modular and can be easily substituted for adapters.

Another major limitation of our work is that we cannot explore whether we can learn different tasks over a given pair of domains. For example, for a given pair of domains such as \textsc{news} and \textsc{twitter}, it would be ideal if we learned a domain adapter and reused it for different applications such as sentiment analysis, named entity recognition, among others. We are limited by the availability of data for such scenarios and this would be a potential future work.

%
\bibliographystyle{acl_natbib}
\bibliography{anthology}
\renewcommand{\arraystretch}{1.2}
\begin{table*}[t!]
\centering \footnotesize 
\resizebox{\linewidth}{!}{%
 \begin{tabular}{l|c|ccccc} 
 \hline
 & \colorbox{red!30}{\textbf{Fully Supervised}} & \multicolumn{5}{c}{\colorbox{blue!25}{\textbf{Adapter Based}}} \\ 
 Src \textrightarrow Trg & \fireemoji & \dannadapter{} & \dsnadapter{} & \taskadapter{} & \ourmethod{} & \jointdt{} \\ [0.5ex]
 \hline
    \apparel{} \textrightarrow{} \baby{} & 87.68 \scriptsize(1.92) & 86.46 \scriptsize(0.26) & 87.13 \scriptsize (0.23) & 87.03 \scriptsize(0.26) & 88.24 \scriptsize(0.76) & \textbf{88.74 \scriptsize (0.13)} \\
    
    \apparel{} \textrightarrow{} \books{} & 83.73 \scriptsize(1.61) & 78.41 \scriptsize(1.14) & 80.23 \scriptsize (0.81) & 84.15 \scriptsize(1.10) & 84.22 \scriptsize(0.76) & \textbf{84.96 \scriptsize (0.28)} \\ 
    
    \apparel{} \textrightarrow{} \cameraphoto{} & 90.00 \scriptsize(1.17) & 87.31 \scriptsize(0.39) & 87.58 \scriptsize (0.48) & \textbf{89.67 \scriptsize(0.32)}  & 88.76 \scriptsize(1.32) & 89.39 \scriptsize (0.23) \\
    
    \apparel{} \textrightarrow{} \moviereviews{} & 76.57 \scriptsize(0.36) & 75.54 \scriptsize(0.63) & 75.96 \scriptsize (0.27) & 76.63 \scriptsize(0.92) & 77.39 \scriptsize(0.13) & \textbf{77.63 \scriptsize (0.71}) \\
    
    \baby{} \textrightarrow{} \apparel{} & 88.56 \scriptsize(1.04) & 87.72 \scriptsize(1.85) & 87.62 \scriptsize (0.86) & 88.33 \scriptsize(1.10) & 89.55 \scriptsize(0.10) & \textbf{89.70 \scriptsize (0.23)} \\
    
    \baby{} \textrightarrow{} \books{} & 85.52 \scriptsize(0.59) & 82.89 \scriptsize(3.08) & 84.26 \scriptsize (0.85) & 84.61 \scriptsize(0.39) & 84.38 \scriptsize(0.61) & \textbf{85.01 \scriptsize (0.60} \\
    
    \baby{} \textrightarrow{} \cameraphoto{} & 89.58 \scriptsize(0.32) & 86.63 \scriptsize(0.53) & 88.44 \scriptsize (0.90) & \textbf{90.63 \scriptsize(0.33)}  & 87.46 \scriptsize(0.88) & 88.64 \scriptsize (0.30) \\
    
    \baby{} \textrightarrow{} \moviereviews{} & 77.26 \scriptsize(0.71) & 74.48 \scriptsize(1.79) & 48.67 \scriptsize (15.98) & 78.74 \scriptsize(0.35) & \textbf{79.42 \scriptsize(0.44)} & 78.44 \scriptsize (0.70)\\
    
    \books{} \textrightarrow{} \apparel{} & 87.38 \scriptsize(1.08) & 85.90 \scriptsize(0.12) & 86.62 \scriptsize (0.41) & 85.03 \scriptsize(0.36)  & 84.79 \scriptsize(0.75) & \textbf{87.46 \scriptsize (0.27)} \\
    
    \books{} \textrightarrow{} \baby{} & 84.72 \scriptsize(1.15) & 82.06 \scriptsize(1.15) & 82.75 \scriptsize (1.51) & 86.50 \scriptsize(0.39)  & \textbf{86.84 \scriptsize(0.48)} & 86.41  \scriptsize (0.79)\\
    
    \books{} \textrightarrow{} \cameraphoto{} & 87.58 \scriptsize(0.67) & 86.94 \scriptsize(0.83) & 86.61 \scriptsize (1.03) & 88.44 \scriptsize(0.53) & 87.86 \scriptsize(0.61) & \textbf{88.53 \scriptsize (0.43)} \\
    
    \books{} \textrightarrow{} \moviereviews{} & 80.14 \scriptsize(0.52) & 76.19 \scriptsize(0.89) & 72.08 \scriptsize (7.29) & 79.44 \scriptsize(0.95)  & \textbf{80.52 \scriptsize(0.61)}  & 78.91 \scriptsize (0.38)\\
    
    \cameraphoto{}{} \textrightarrow{} \apparel{} & 89.46 \scriptsize(0.49) & 87.02 \scriptsize(1.86)& 85.50 \scriptsize (1.30) & 87.74 \scriptsize(1.18)  & 88.53 \scriptsize{(0.42)} & \textbf{88.92 \scriptsize (0.44)} \\
    
    \cameraphoto{} \textrightarrow{} \baby{} & 90.15 \scriptsize(0.46) & 88.10 \scriptsize(1.13) & 88.56 \scriptsize (0.25) & 81.71 \scriptsize(2.72)  & \textbf{89.72 \scriptsize(0.43)} & 89.32 \scriptsize (0.42) \\
    
    \cameraphoto{} \textrightarrow{} \books{} & 85.08 \scriptsize(0.97) & 81.18 \scriptsize(2.07) & 83.81 \scriptsize (1.68) & 80.55 \scriptsize(0.81) & 84.14 \scriptsize(0.52) & \textbf{85.42 \scriptsize (0.70)} \\
    
    \cameraphoto{} \textrightarrow{} \moviereviews{} &  76.03 \scriptsize(1.15) & 64.99 \scriptsize(5.91) & 63.59 \scriptsize (11.98) & 69.53 \scriptsize(1.24)  & 73.22 \scriptsize(0.48) & \textbf{73.50 \scriptsize (0.84)} \\
    
    \moviereviews{} \textrightarrow{} \apparel{} & 79.55 \scriptsize(1.38) & 81.05 \scriptsize(1.15) & 66.28 \scriptsize (19.68) & 82.45 \scriptsize(1.43) & 81.93 \scriptsize(0.47) & \textbf{84.41 \scriptsize (0.43)} \\
    
    \moviereviews{} \textrightarrow{} \baby{} & 74.63 \scriptsize(9.8) & 77.95 \scriptsize(1.46) & 54.64 \scriptsize (17.71) & 81.70 \scriptsize(1.22) & 84.28 \scriptsize(0.41) & \textbf{84.91 \scriptsize (0.36)} \\
    
    \moviereviews{} \textrightarrow{} \books{} & 86.09 \scriptsize(1.0) & 82.83 \scriptsize(0.62) & 49.92 \scriptsize(24.06) & \textbf{84.90 \scriptsize (0.23)}  & 84.47 \scriptsize(0.80) & 84.45 \scriptsize (0.31) \\
    
    \moviereviews{} \textrightarrow{} \cameraphoto{} & 76.54 \scriptsize(1.78) & 84.58 \scriptsize(0.46) & 69.47 \scriptsize (12.49) & 86.68 \scriptsize(0.65)  & 86.25 \scriptsize(0.38) & \textbf{88.37 \scriptsize (0.11)} \\  [0.5ex] \hline
    
    Avg & 83.81 \scriptsize(1.41) & 81.91 \scriptsize (1.37)  & 76.49 \scriptsize(5.98) & 83.72 \scriptsize(0.88)  & 84.60 \scriptsize(0.57) & \textbf{85.16 \scriptsize (0.43)} \\   
 \hline
 \end{tabular} %
 }
 \caption{\label{tab:amazon_results_acl}F1 scores for \amazondataset{} dataset. We report the mean and standard deviation of 3 runs. The five domains are Apparel (\apparel{}), Baby (\baby{}), Books (\books{}), Camera\_Photo (\cameraphoto{}) and Movie Reviews (\moviereviews{}). The difference between this table and \Cref{tab:amazon_results} is we experiment with \dsnadapter{}. \fireemoji fine-tunes a language model using labeled data from the source domain and tests it on the target domain. This shows that just using the supervised data from the source domain is not enough}
\end{table*}

\renewcommand{\arraystretch}{1.2}
\begin{table*}[t!]
\centering \footnotesize 
\resizebox{\linewidth}{!}{%
 \begin{tabular}{l|c|ccccc} 
 \hline
 & \colorbox{red!30}{\textbf{Fully Supervised}} & \multicolumn{5}{c}{\colorbox{blue!25}{\textbf{Adapter Based}}} \\ 
 Src \textrightarrow Trg & \fireemoji & \dannadapter{} & \dsnadapter{} & \taskadapter{} & \ourmethod{} & \jointdt{} \\ [0.5ex] 
 \hline
    \fiction{} \textrightarrow \slate{} & 71.58 \scriptsize(0.31) & 70.96 \scriptsize(0.03) & 70.16 \scriptsize(0.25) & 72.36 \scriptsize(0.36) & \textbf{73.46 \scriptsize(0.34)} & 72.30 \scriptsize(0.26)  \\
    
    \fiction{} \textrightarrow \government{} & 79.05 \scriptsize(0.94) & 78.73 \scriptsize(0.43) & 77.01 \scriptsize(0.31) & 79.00 \scriptsize(0.46) & 78.65 \scriptsize(0.25) & \textbf{79.79 \scriptsize(0.22)} \\
    
    \fiction{} \textrightarrow \telephone{} & 74.73 \scriptsize(0.41) & 70.89 \scriptsize(0.74) & 69.89 \scriptsize(0.04) & 70.83 \scriptsize(0.54) & \textbf{73.05 \scriptsize(0.70)} & 71.59 \scriptsize(0.78) \\
    
    \fiction \textrightarrow \travel{} & 75.84 \scriptsize(0.48) & 74.42 \scriptsize(0.18) & 73.98 \scriptsize(0.70) & 75.85 \scriptsize(0.19) & 76.75 \scriptsize(0.80) & \textbf{77.07 \scriptsize(0.26)} \\
    
    \slate{} \textrightarrow \fiction{} & 76.27 \scriptsize(0.30) & 73.89 \scriptsize(0.61) & 73.79 \scriptsize(0.06) & 75.25 \scriptsize(0.19) & \textbf{75.52 \scriptsize(0.89)} & 75.35 \scriptsize(0.56) \\
    
    \slate{} \textrightarrow \government{} & 81.00 \scriptsize(0.37) & 79.99 \scriptsize(0.36) & 79.39 \scriptsize(0.16) & 80.76 \scriptsize(0.40) & \textbf{81.65 \scriptsize(0.11)} & 80.94 \scriptsize(0.30) \\
    
    \slate{} \textrightarrow \telephone{} & 74.32 \scriptsize(0.71) & 72.29 \scriptsize(0.57) & 71.69 \scriptsize(0.16) & 72.66 \scriptsize(0.79) & \textbf{74.09 \scriptsize(0.30)} & 73.38 \scriptsize(0.63) \\
    
    \slate{} \textrightarrow \travel{} & 77.85 \scriptsize(0.40) & 75.58 \scriptsize(0.54) & 75.24 \scriptsize(0.42) & 76.16 \scriptsize(0.22) & \textbf{77.31 \scriptsize(0.60)} & 77.16 \scriptsize(0.18) \\
    
    \government{} \textrightarrow \fiction{} & 73.12 \scriptsize(0.39) & 71.57 \scriptsize(0.68) & 70.67 \scriptsize(0.29) & 72.66 \scriptsize(0.31) & 72.66 \scriptsize(0.56) & \textbf{73.56 \scriptsize(0.23)} \\
    
    \government{} \textrightarrow \slate{} & 72.10 \scriptsize(1.01) & 70.17 \scriptsize(0.64) & 70.31 \scriptsize(0.44) & 71.11 \scriptsize(0.38) & 71.14 \scriptsize(0.21) & \textbf{71.36 \scriptsize(0.04)} \\
    
    \government{} \textrightarrow \telephone{} & 72.80 \scriptsize(0.32) & 69.45 \scriptsize(0.96) & 69.47 \scriptsize(0.25) & 71.40 \scriptsize(0.30) & 71.53 \scriptsize(1.04) & \textbf{71.99 \scriptsize(0.67)} \\
    
    \government{} \textrightarrow \travel{} & 76.76 \scriptsize(0.08) & 74.35 \scriptsize(0.22) & 74.00 \scriptsize(0.32) & 76.29 \scriptsize(0.10) & 76.16 \scriptsize(0.34) & \textbf{76.79 \scriptsize(0.59)} \\
    
    \telephone{} \textrightarrow \fiction{} & 73.25 \scriptsize(0.36) & 72.24 \scriptsize(0.59) & 73.04 \scriptsize(0.28) & \textbf{74.48 \scriptsize(0.33)} & 73.34 \scriptsize(0.41) & 73.89 \scriptsize(0.12) \\
    
    \telephone{} \textrightarrow \slate{} & 69.52 \scriptsize(1.17) & 69.09 \scriptsize(1.79) & 69.40 \scriptsize(0.42) & 70.94 \scriptsize(0.16) & 70.94 \scriptsize(0.55) & \textbf{71.41 \scriptsize(0.19)} \\
    
    \telephone{} \textrightarrow \government{} & 77.59 \scriptsize(1.38)  & 77.80 \scriptsize(0.27) & 77.56 \scriptsize(0.46) & 79.24 \scriptsize(0.35) & 79.65 \scriptsize(0.60) & \textbf{79.78 \scriptsize(0.64)} \\
    
    \telephone{} \textrightarrow \travel{} & 72.45 \scriptsize(2.44)  & 74.67 \scriptsize(0.50) & 74.14 \scriptsize(0.21) & 75.27 \scriptsize(0.83) & \textbf{76.11 \scriptsize(0.91)} & 75.95 \scriptsize(0.50) \\
    
    \travel{} \textrightarrow \fiction{} & 72.78 \scriptsize(0.37) & 70.27 \scriptsize(0.45) & 71.10 \scriptsize(0.21) & 72.20 \scriptsize(0.49) & 73.12 \scriptsize(0.08) & \textbf{73.13 \scriptsize(0.22)} \\
    
    \travel{} \textrightarrow \slate{} & 70.40 \scriptsize(0.10) & 68.35 \scriptsize(0.62) & 69.92 \scriptsize(0.50) & 70.28 \scriptsize(0.37) & 70.67 \scriptsize(0.50) & \textbf{71.28 \scriptsize(0.38)} \\
    
    \travel{} \textrightarrow \government{} & 79.75 \scriptsize(0.42) & 79.25 \scriptsize(0.34) & 79.75 \scriptsize(0.24) & 81.26 \scriptsize(0.37) & 81.11 \scriptsize(0.42) & \textbf{81.55 \scriptsize(0.16)} \\
    
    \travel{} \textrightarrow \telephone{} & 72.02 \scriptsize(0.49) & 69.33 \scriptsize(0.41) & 70.10 \scriptsize(0.52) & 70.98 \scriptsize(0.11) & 70.95 \scriptsize(0.19) & \textbf{71.42 \scriptsize(0.12)} \\[0.5ex] \hline
    
    Avg & 74.66 \scriptsize(0.62) & 73.16 \scriptsize(0.55) & 73.03 \scriptsize(0.32) & 74.45 \scriptsize(0.40) & 74.89 \scriptsize(0.49) & \textbf{74.98 \scriptsize(0.35)} \\    
 \hline
 \end{tabular} %
 }
 \caption{\label{tab:mnli_results_acl}F1 scores for \mnlidataset{}. We report mean and standard deviation of 3 runs. The five domains are Fiction (\fiction{}), Slate (\slate{}), Government (\government{}), Telephone (\telephone{}), and Travel (\travel{}). The difference between this table and \Cref{tab:mnli_results} is we experiment with \dsnadapter{}. \fireemoji fine-tunes a language model using labeled data from the source domain and tests it on the target domain. This shows that just using the supervised data from the source domain is not enough.}
\end{table*}

\begin{figure*}[t!]
\centering \footnotesize 
\resizebox{\linewidth}{!}{%
 \begin{tabular}{c@{}c@{}c@{}c}
 \includegraphics[width=\textwidth]{figures/pretrained_slate_travel/layer_1.png}&
     \includegraphics[width=\textwidth]{figures/pretrained_slate_travel/layer_2.png}&
     \includegraphics[width=\textwidth]{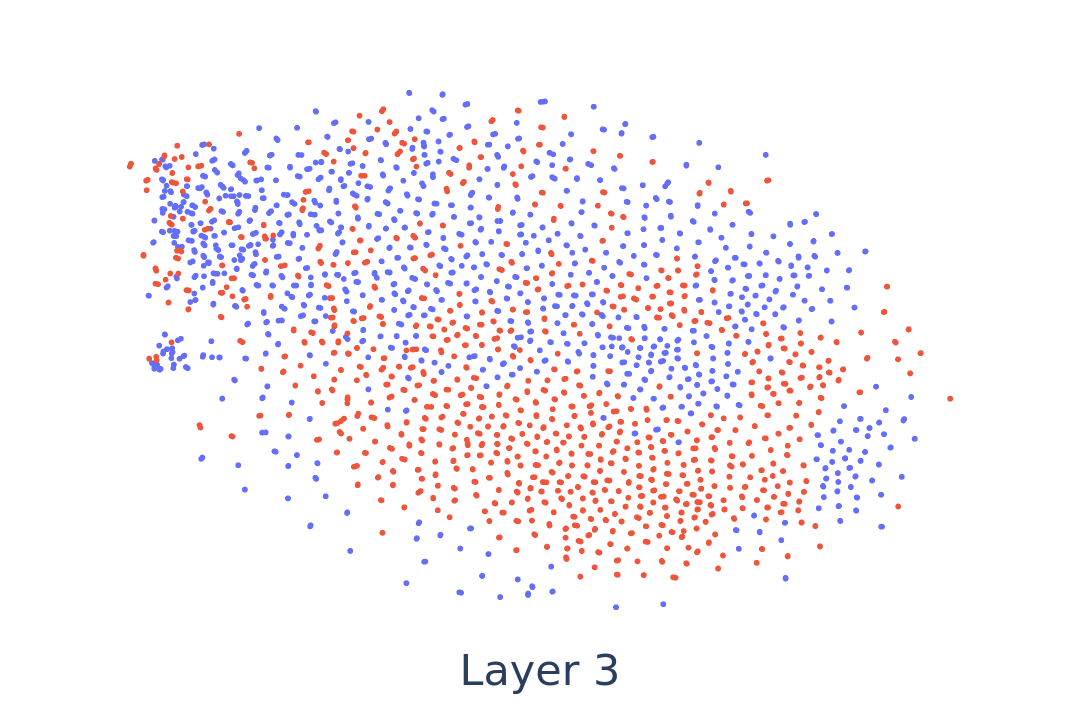}&
     \includegraphics[width=\textwidth]{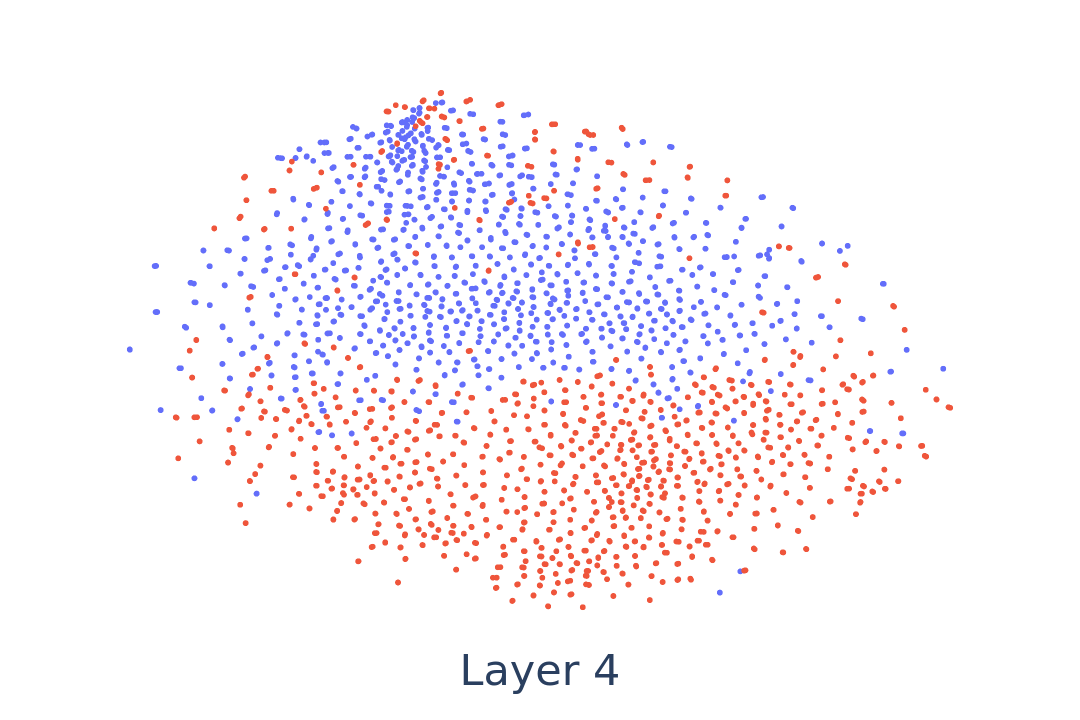} \\
 \includegraphics[width=\textwidth]{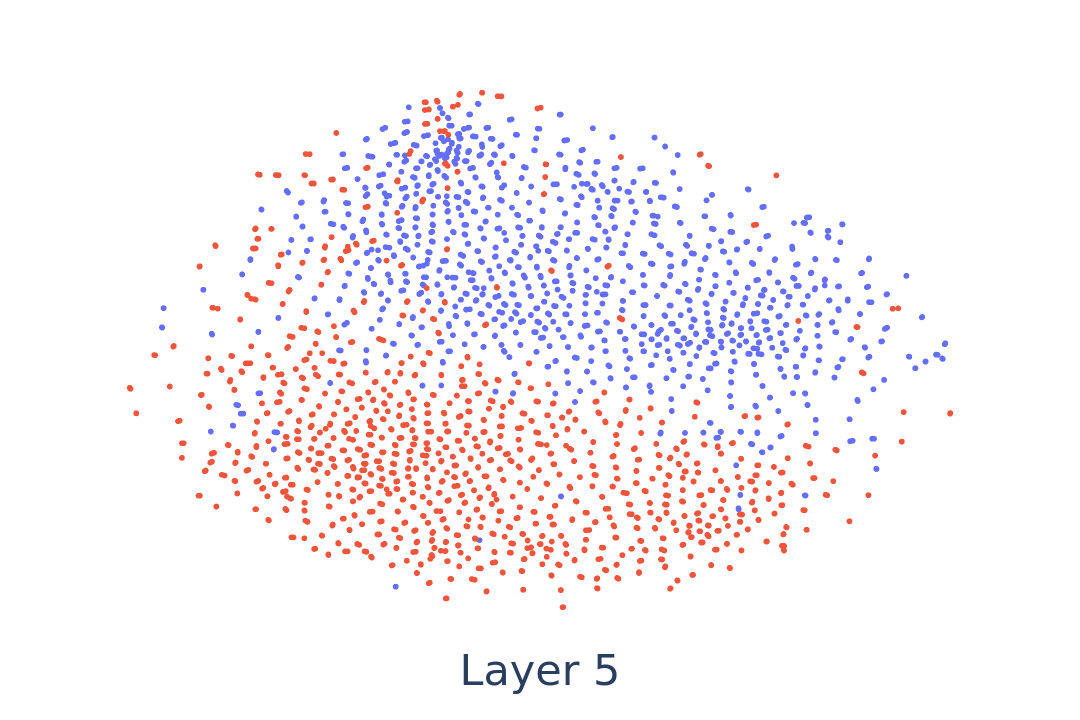}&
     \includegraphics[width=\textwidth]{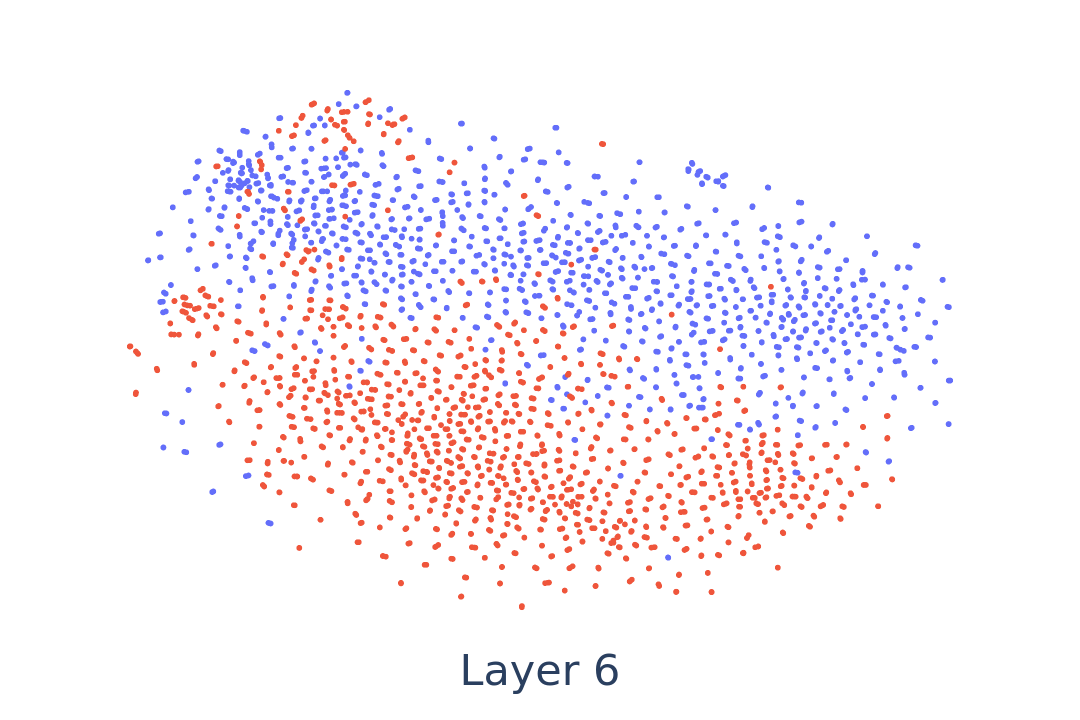}&
     \includegraphics[width=\textwidth]{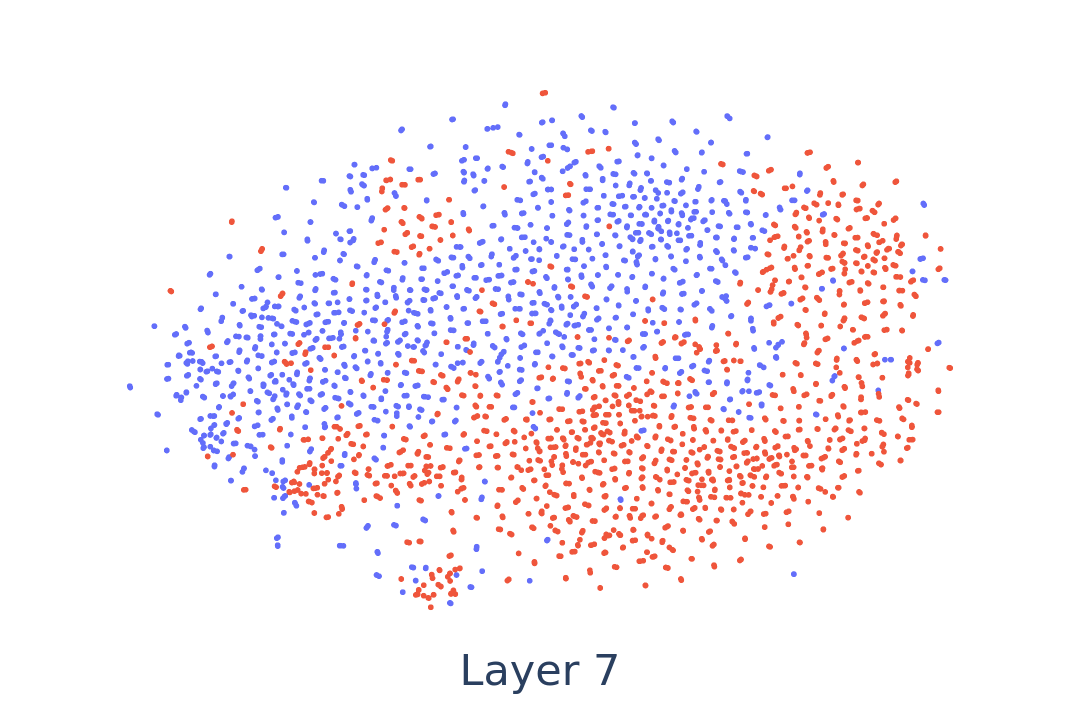}&
     \includegraphics[width=\textwidth]{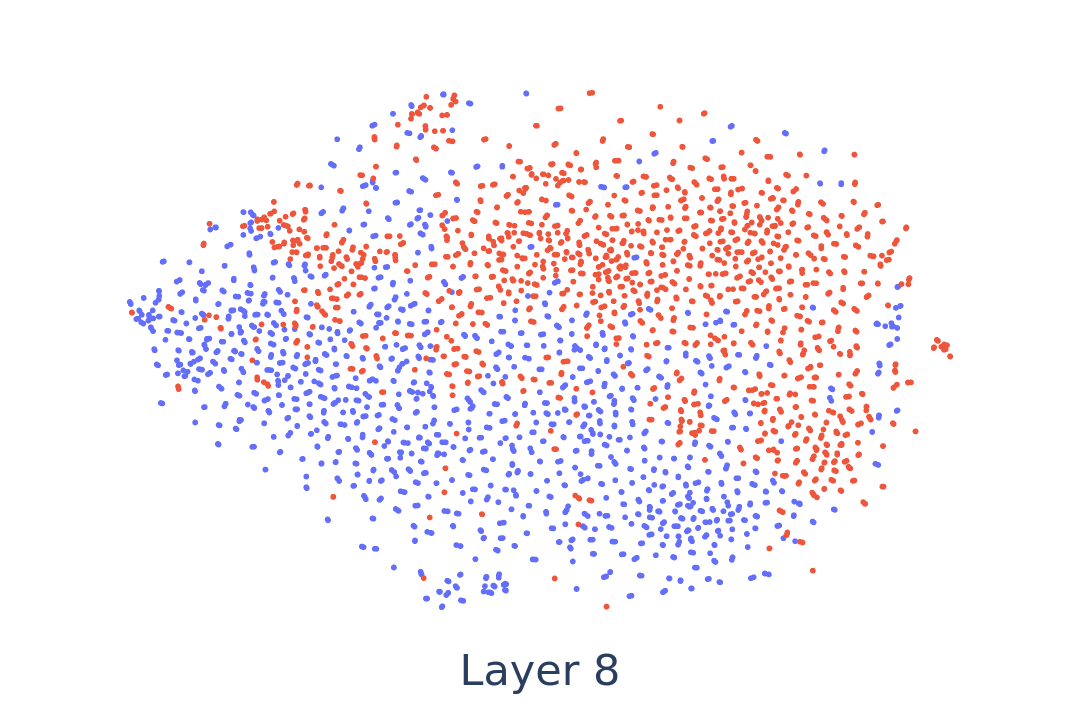} \\
 \includegraphics[width=\textwidth]{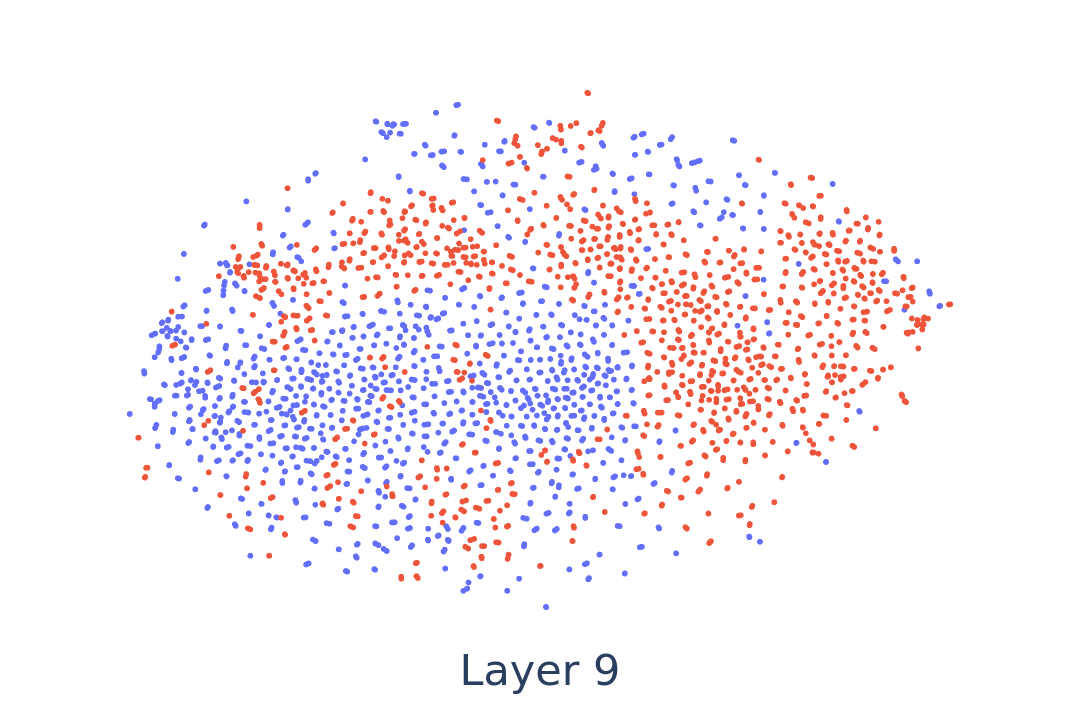}&
     \includegraphics[width=\textwidth]{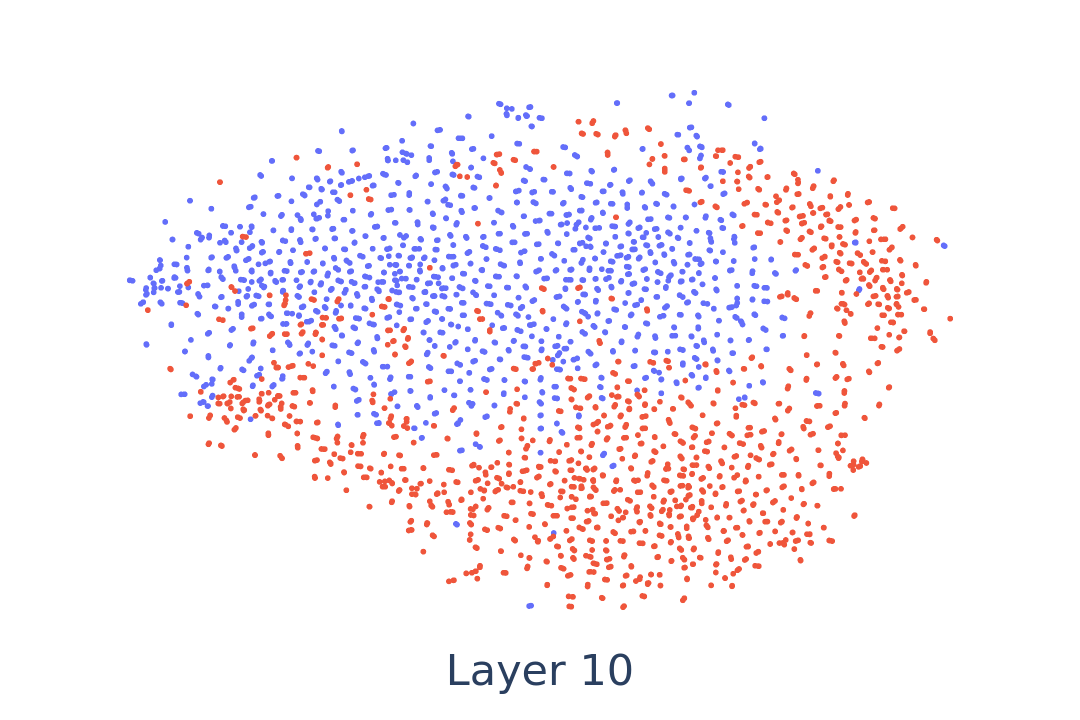}&
     \includegraphics[width=\textwidth]{figures/pretrained_slate_travel/layer_11.png}&
     \includegraphics[width=\textwidth]{figures/pretrained_slate_travel/layer_12.png} \\
  \end{tabular} %
  }
 \caption{t-SNE plots for the pretrained representations from \bertbase{} for \mnlidataset{}. Lower layers are domain-invariant whereas higher layers are domain variant.}
 \label{fig:appendix-mnli-bert-tsne}
\end{figure*}

\begin{figure*}[t!]
\centering \footnotesize 
\resizebox{\linewidth}{!}{%
 \begin{tabular}{c@{}c@{}c@{}c}
 \includegraphics[width=\textwidth]{figures/domain_adapter_slate_travel/layer_1.png}&
     \includegraphics[width=\textwidth]{figures/domain_adapter_slate_travel/layer_2.png}&
     \includegraphics[width=\textwidth]{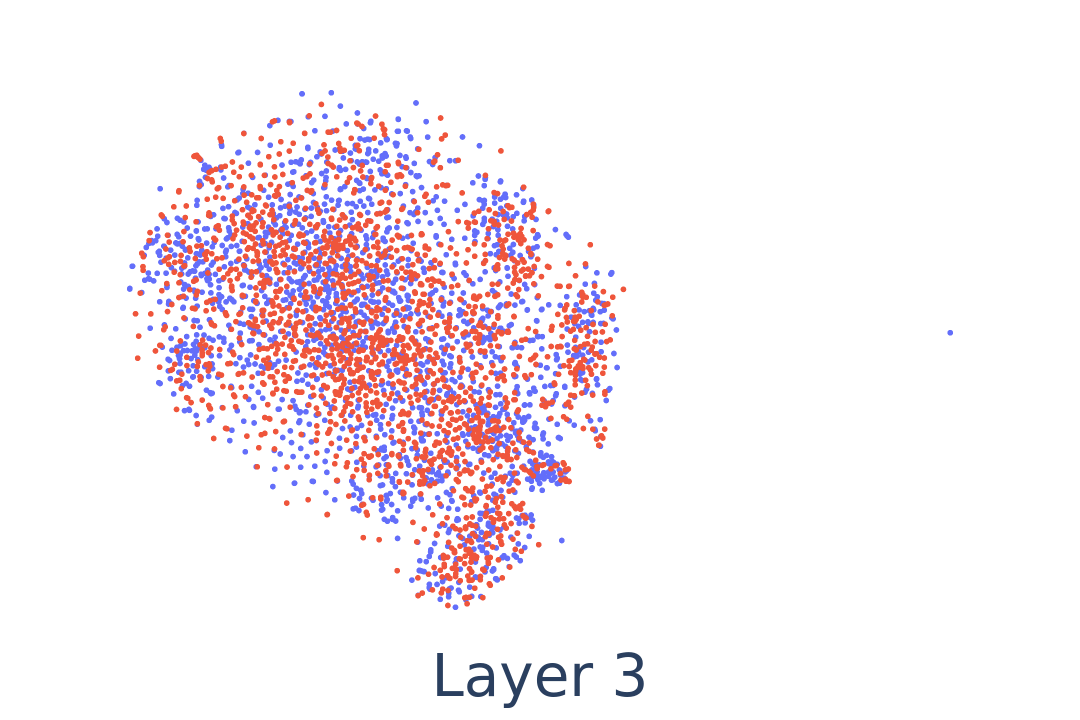}&
     \includegraphics[width=\textwidth]{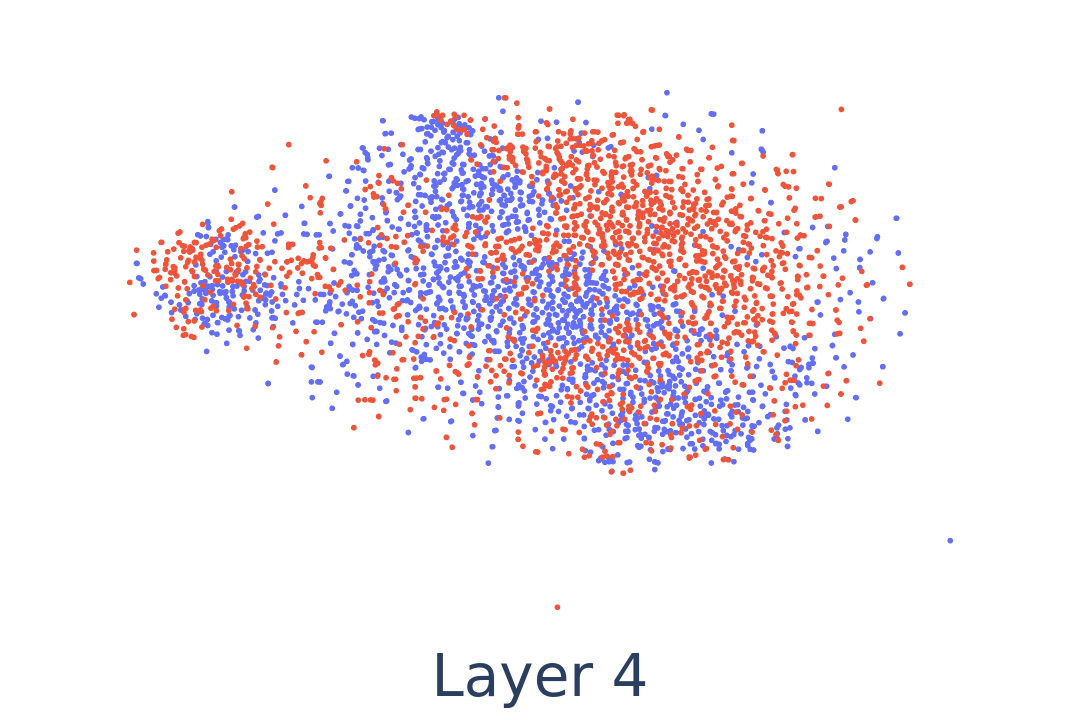} \\
 \includegraphics[width=\textwidth]{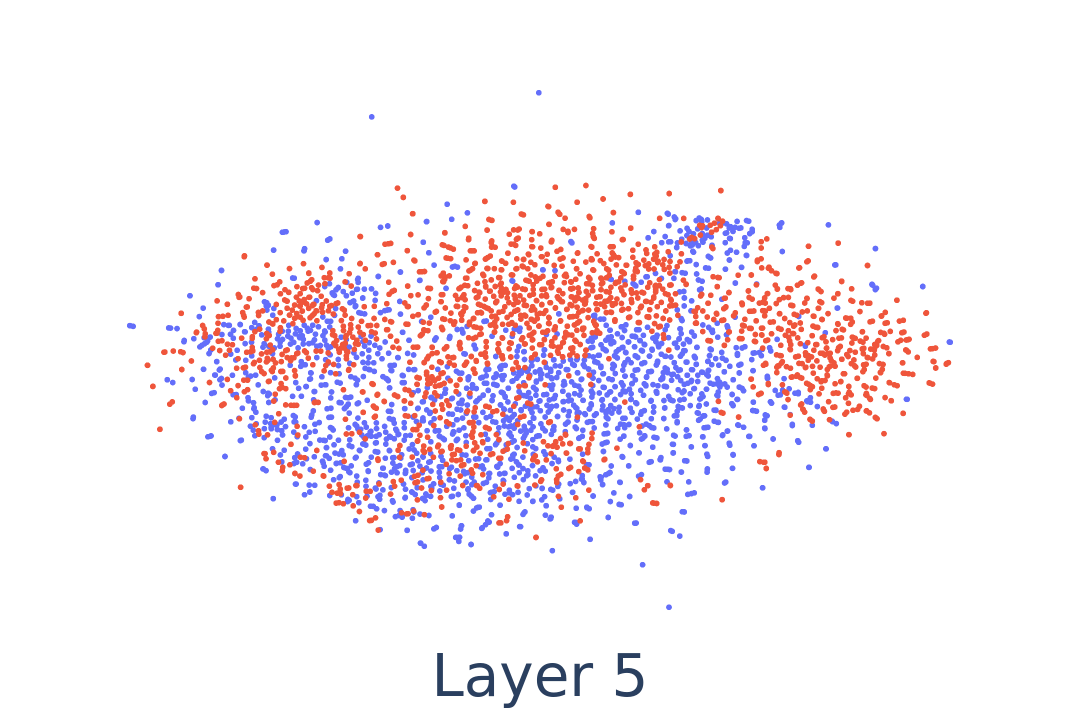}&
     \includegraphics[width=\textwidth]{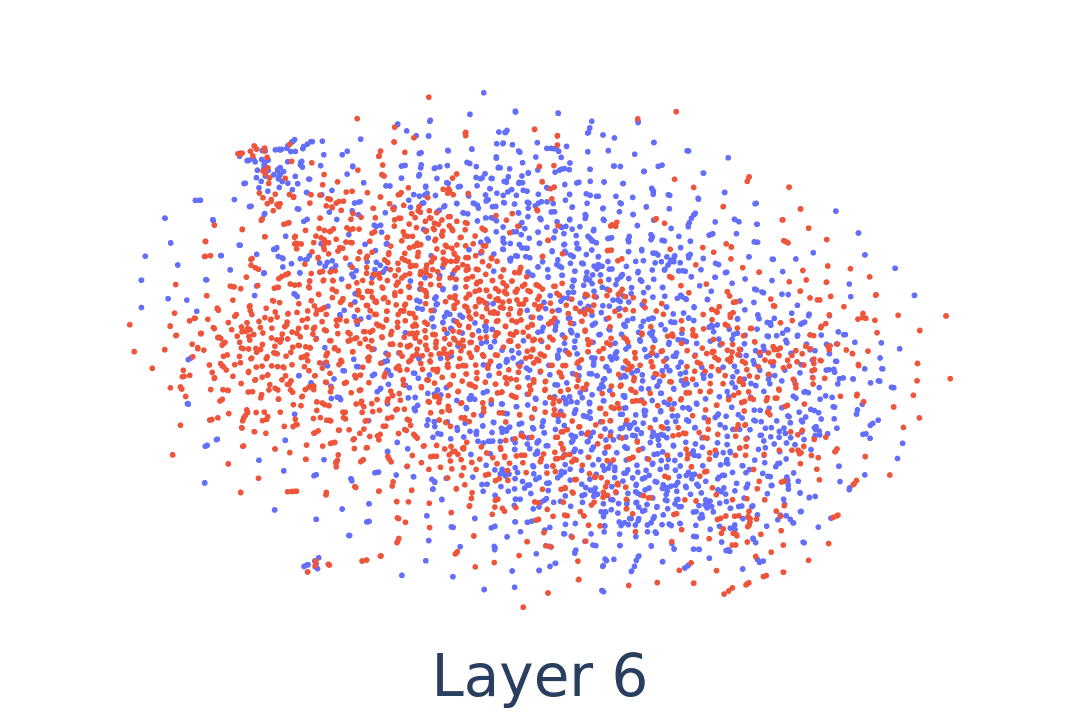}&
     \includegraphics[width=\textwidth]{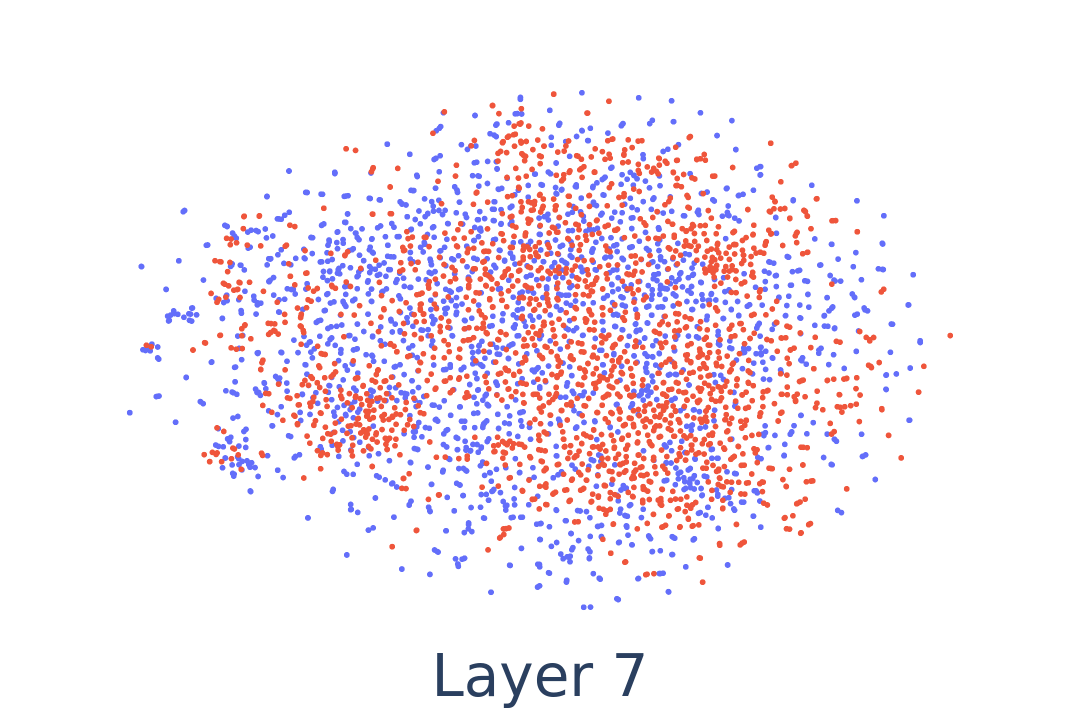}&
     \includegraphics[width=\textwidth]{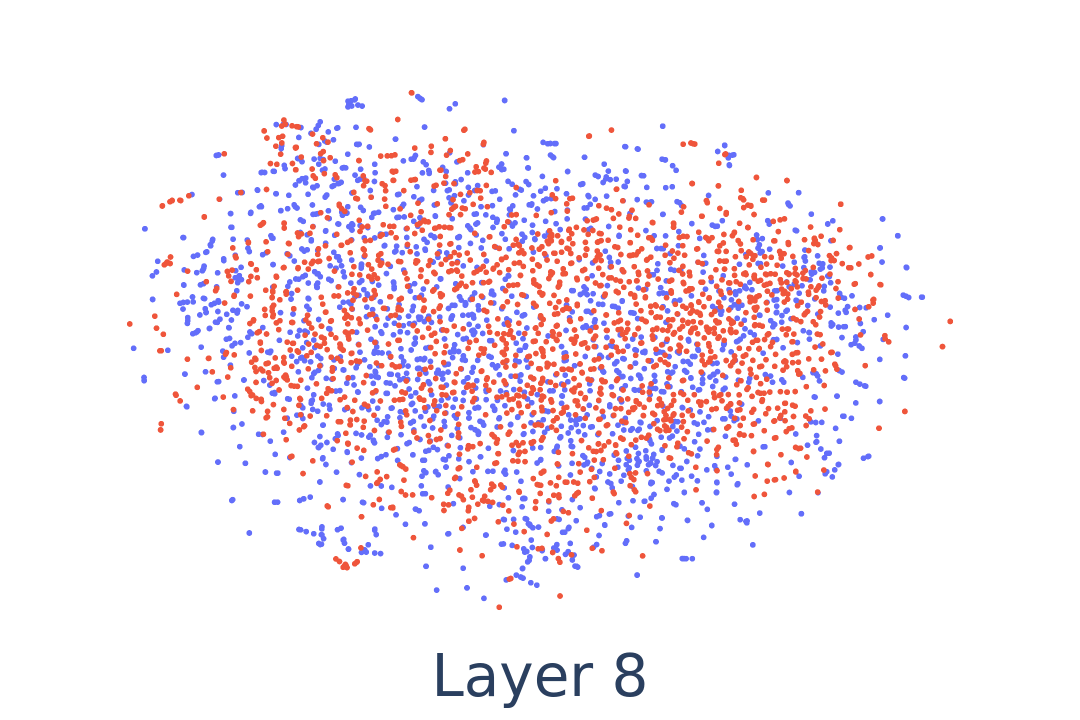} \\
 \includegraphics[width=\textwidth]{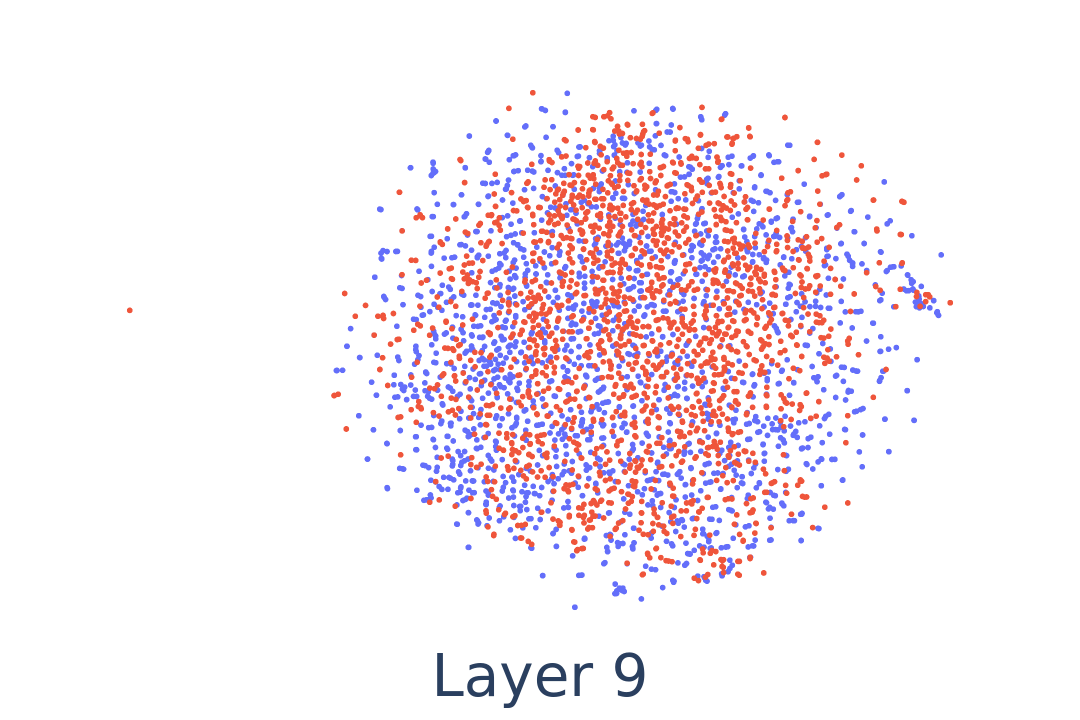}&
     \includegraphics[width=\textwidth]{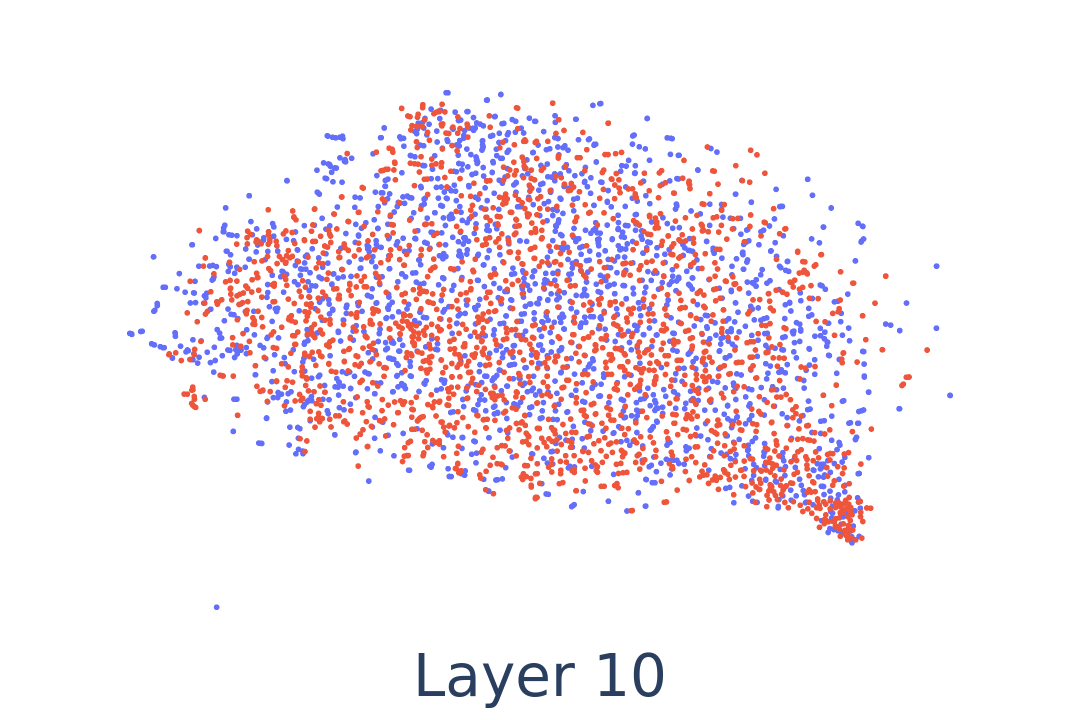}&
     \includegraphics[width=\textwidth]{figures/domain_adapter_slate_travel/layer_11.png}&
     \includegraphics[width=\textwidth]{figures/domain_adapter_slate_travel/layer_12.png} \\
  \end{tabular} %
  }
 \caption{t-SNE plots for the representations from domain adapter trained on \slate{} \textrightarrow{} \travel{} domain for \mnlidataset{}. We reduce divergence between domains for all layers.}
 \label{fig:appendix-mnli-domain-adapter-tsne}
\end{figure*}

\begin{figure*}[t!]
\centering \footnotesize 
\resizebox{\linewidth}{!}{%
 \begin{tabular}{c@{}c@{}c@{}c}
 \includegraphics[width=\textwidth]{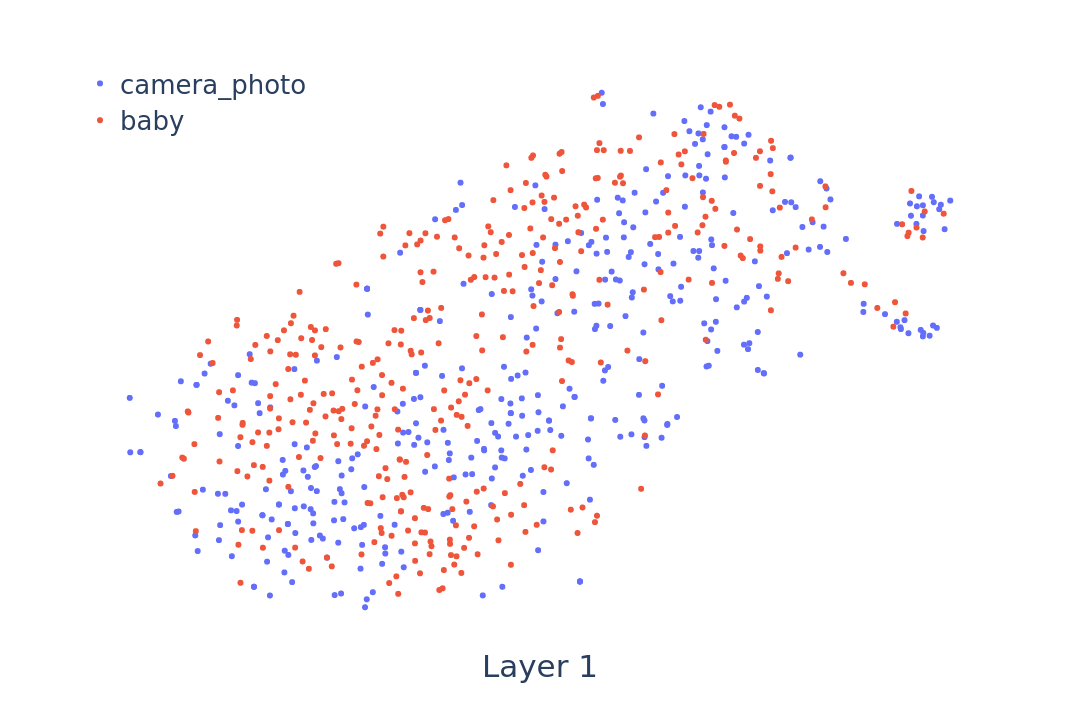}&
     \includegraphics[width=\textwidth]{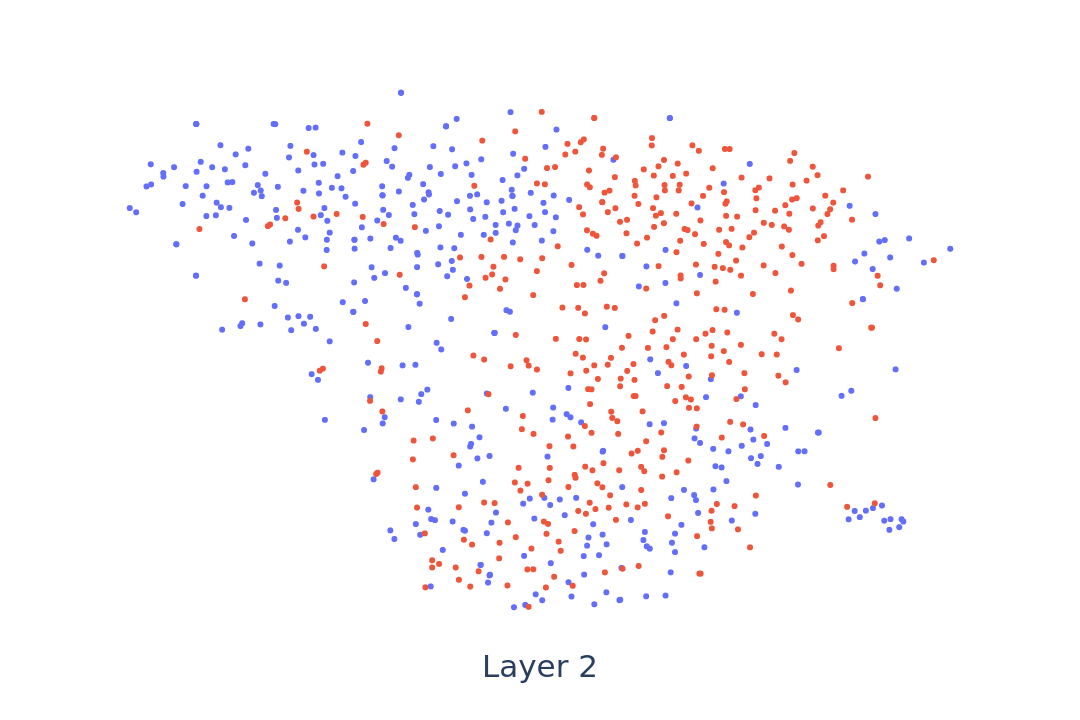}&
     \includegraphics[width=\textwidth]{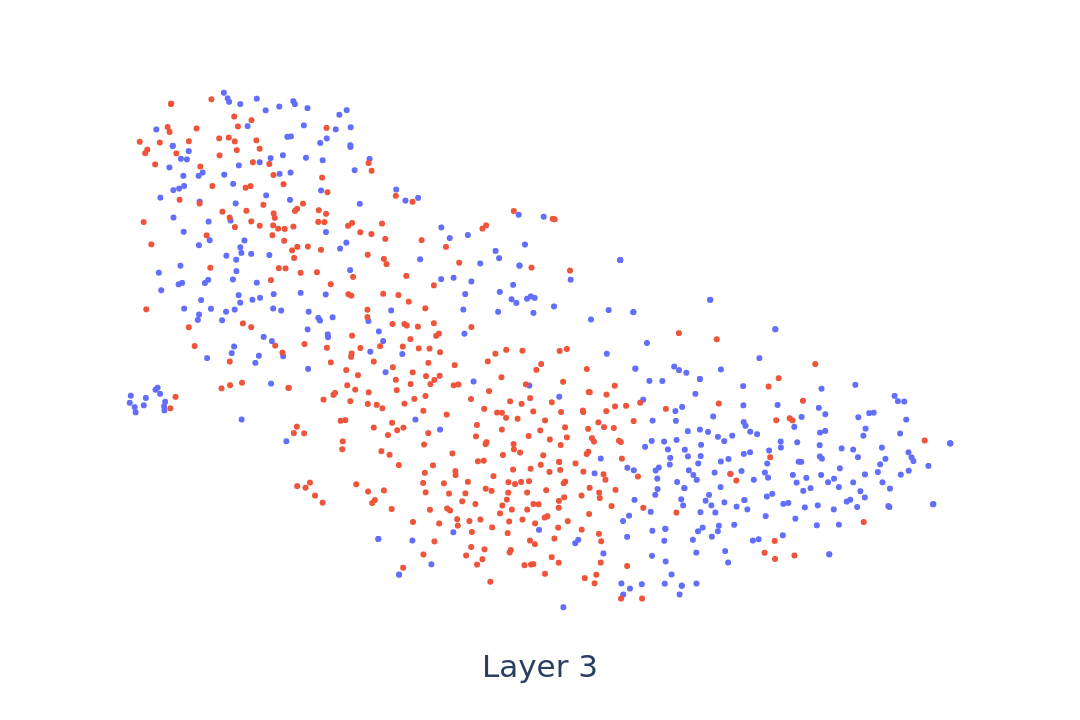}&
     \includegraphics[width=\textwidth]{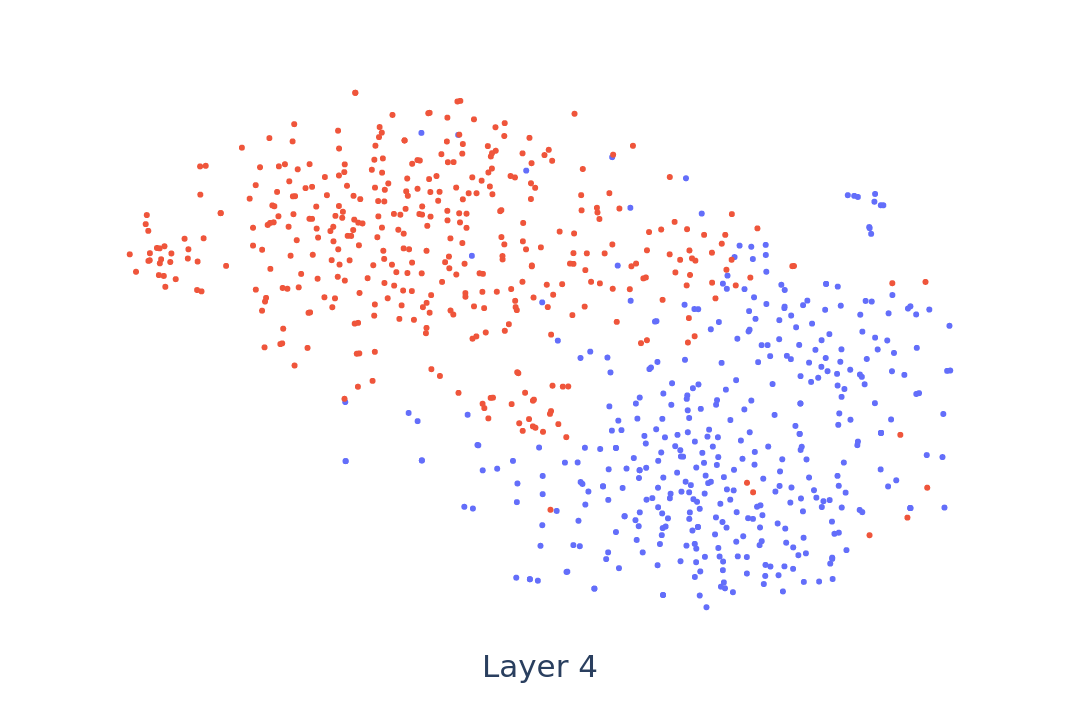} \\
 \includegraphics[width=\textwidth]{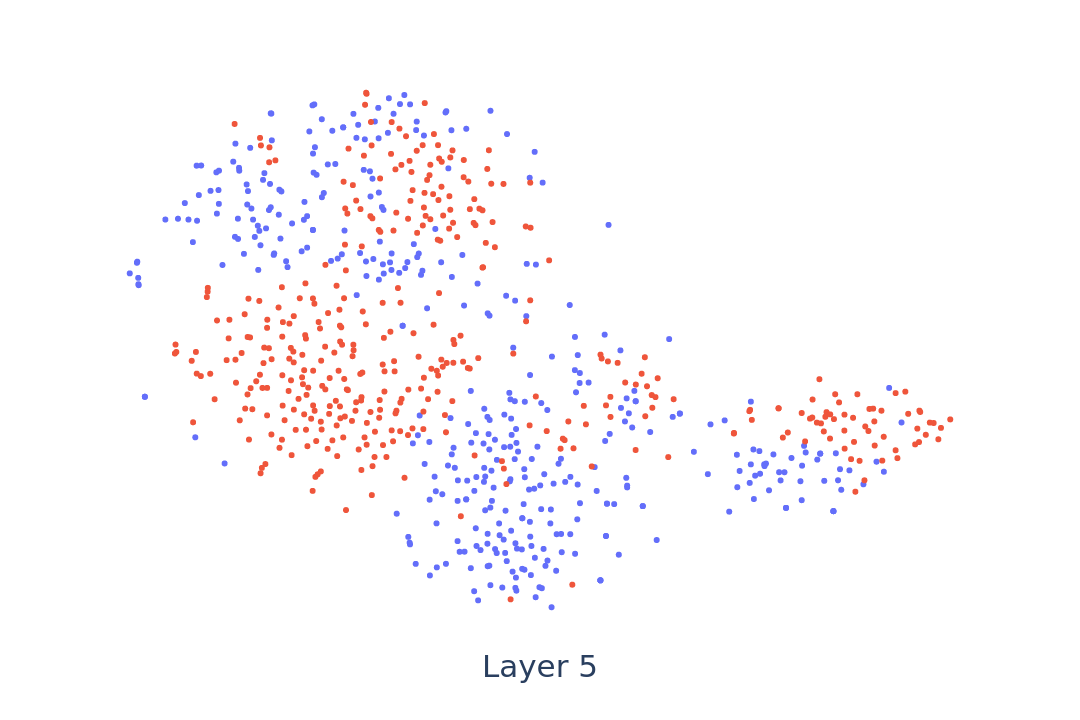}&
     \includegraphics[width=\textwidth]{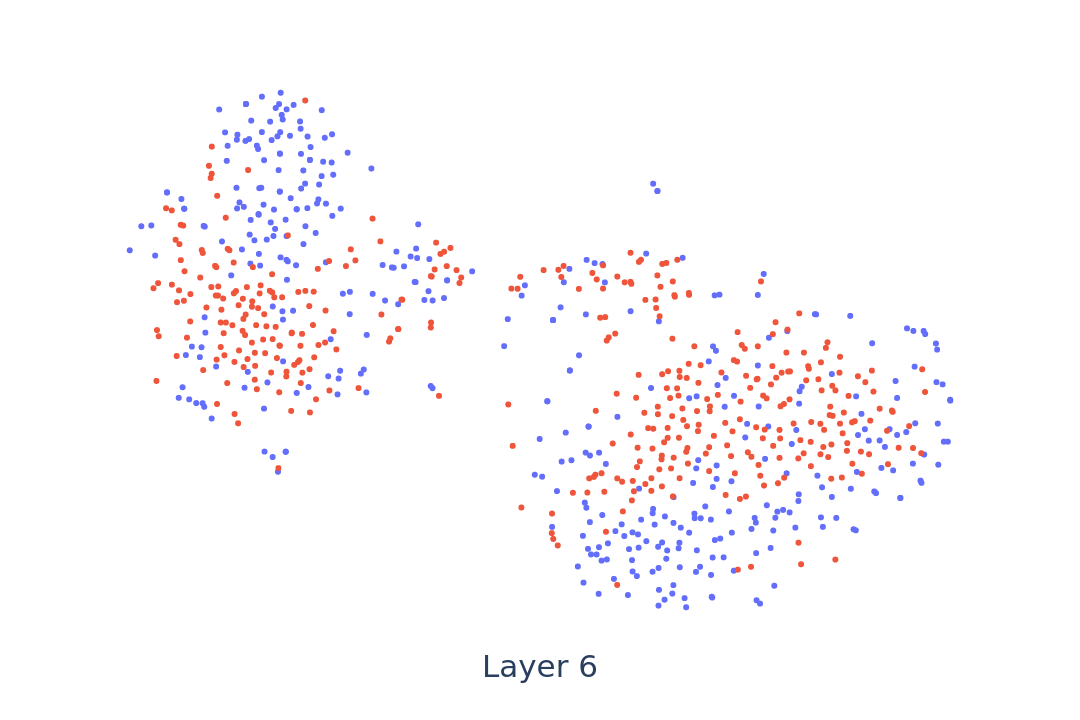}&
     \includegraphics[width=\textwidth]{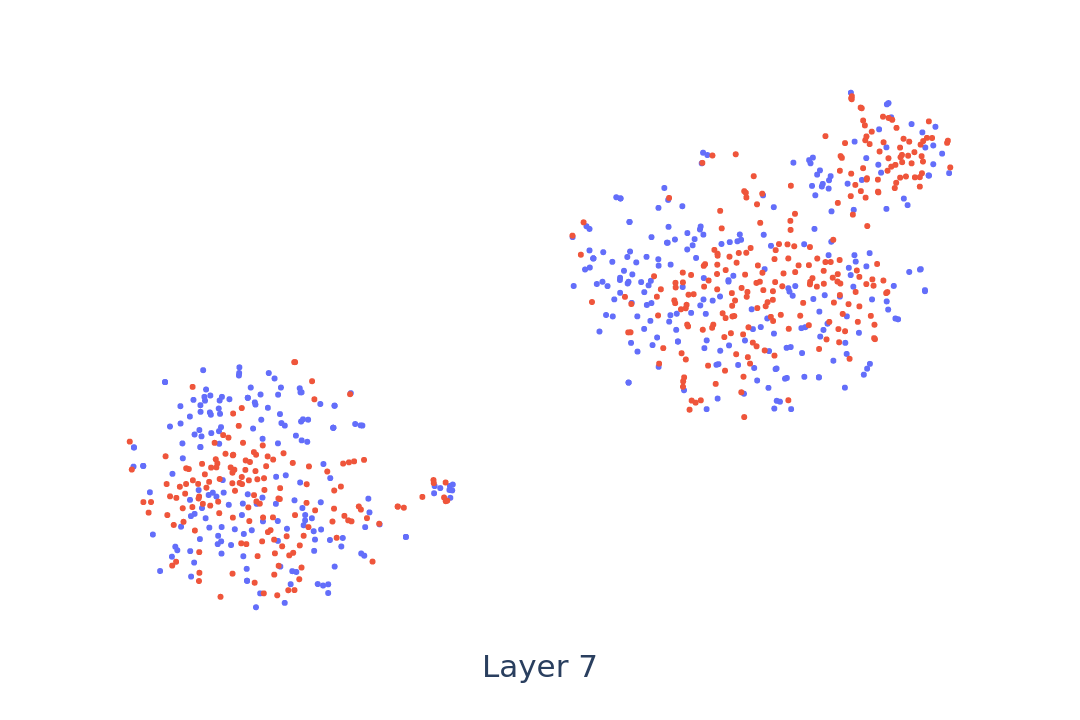}&
     \includegraphics[width=\textwidth]{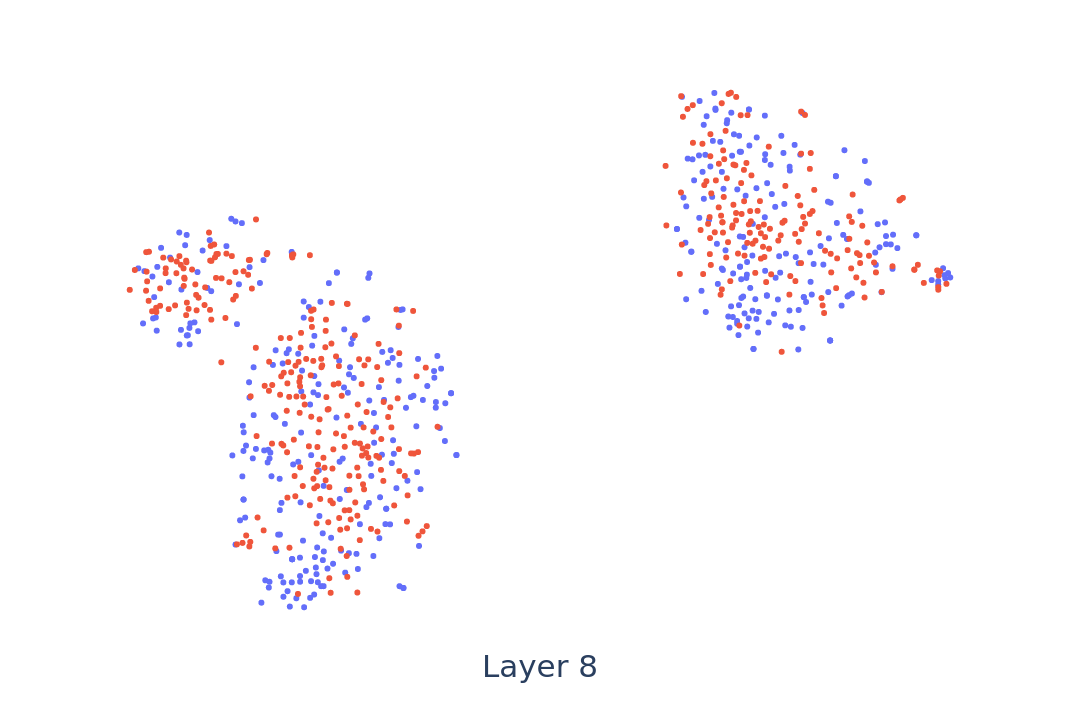} \\
 \includegraphics[width=\textwidth]{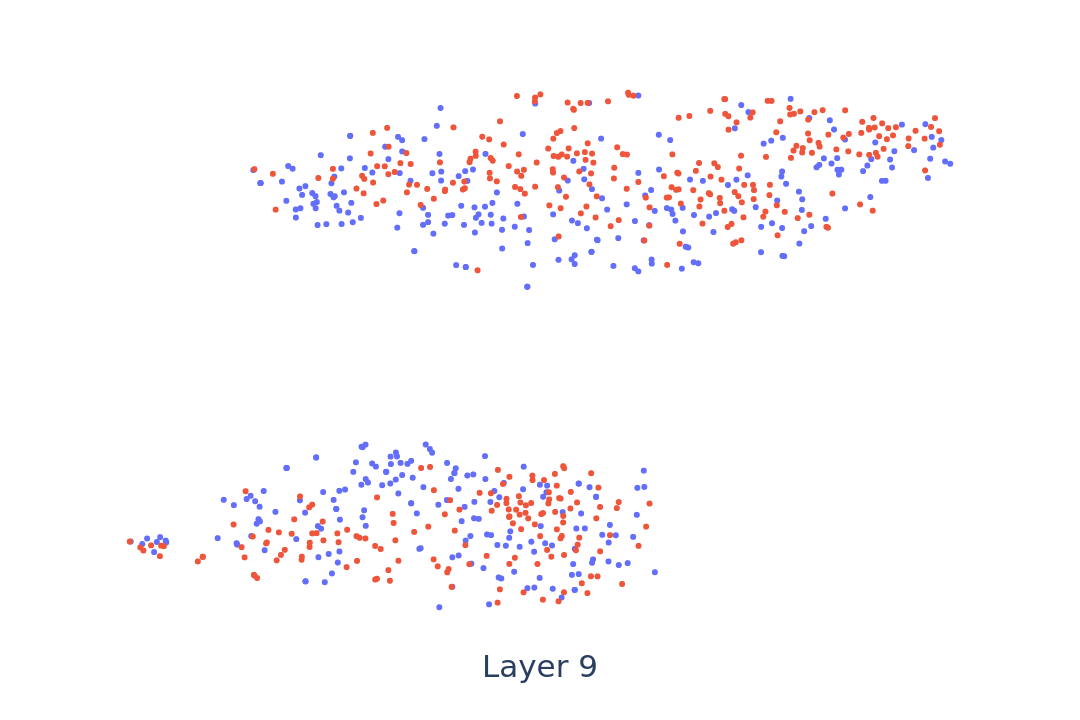}&
     \includegraphics[width=\textwidth]{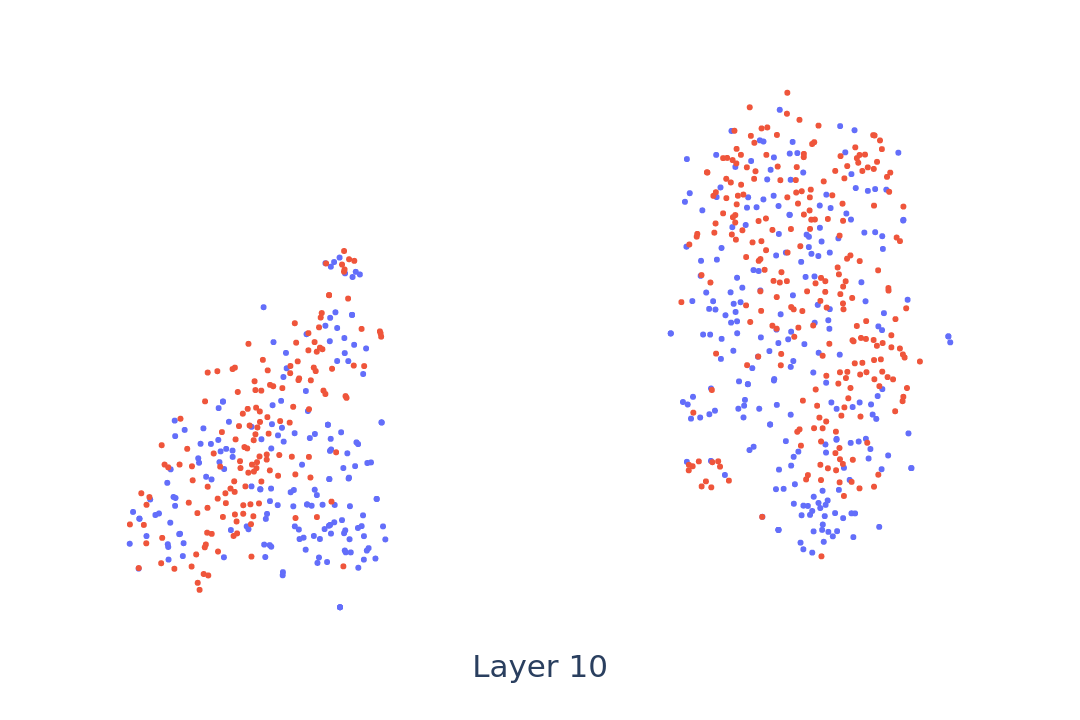}&
     \includegraphics[width=\textwidth]{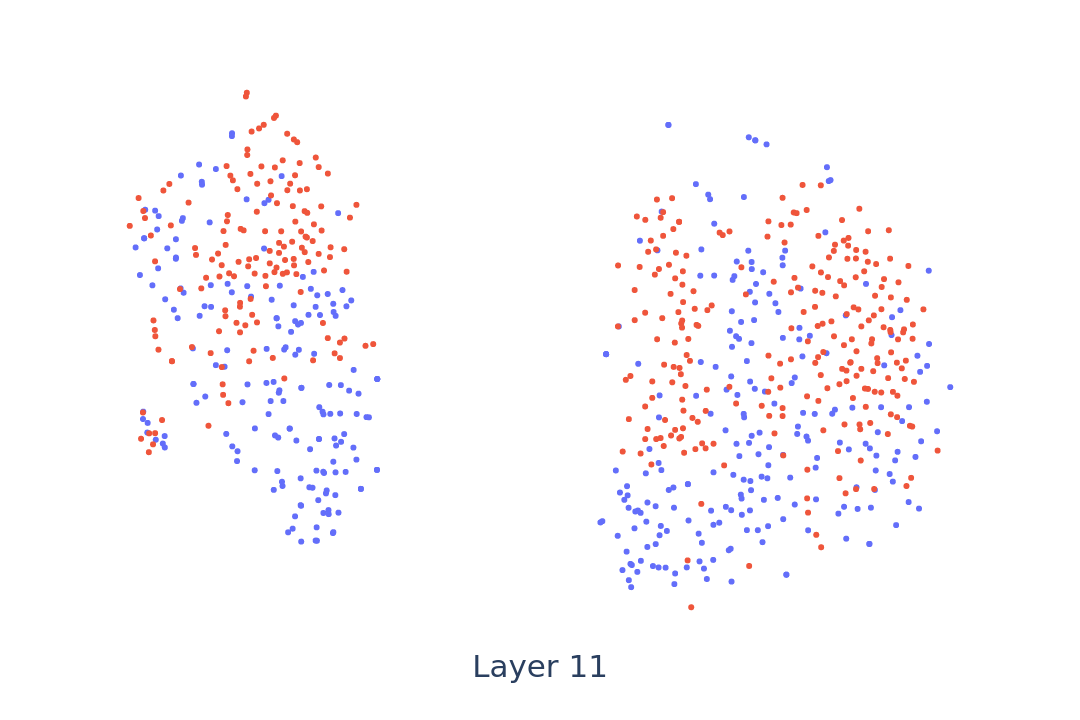}&
     \includegraphics[width=\textwidth]{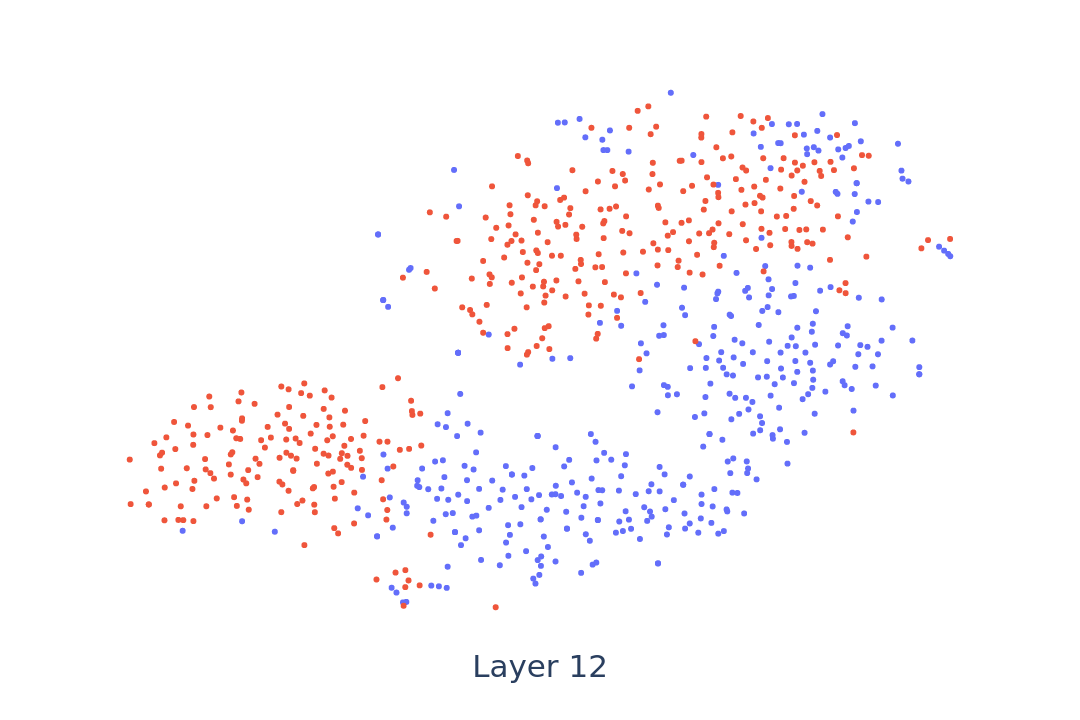} \\
  \end{tabular} %
  }
 \caption{t-SNE plots for the pretrained representations from \bertbase{} for \amazondataset{}. Lower layers are domain-invariant whereas higher layers are domain variant.}
 \label{fig:appendix-amazon-bert-tsne}
\end{figure*}

\begin{figure*}[t!]
\centering \footnotesize 
\resizebox{\linewidth}{!}{%
 \begin{tabular}{c@{}c@{}c@{}c}
 \includegraphics[width=\textwidth]{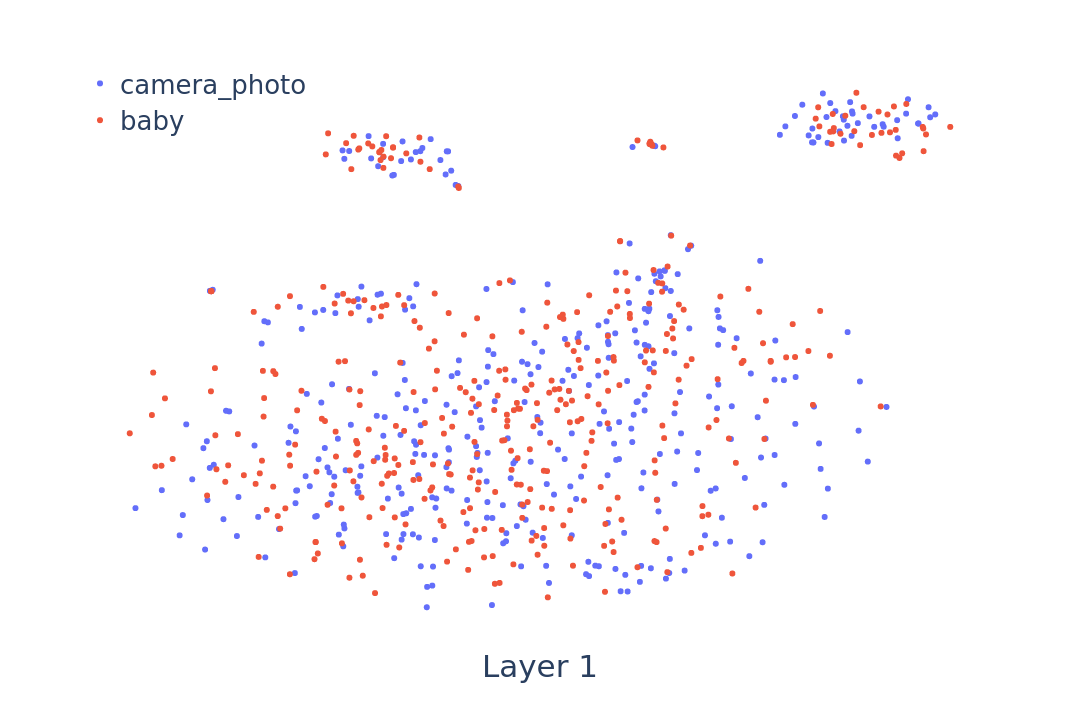}&
     \includegraphics[width=\textwidth]{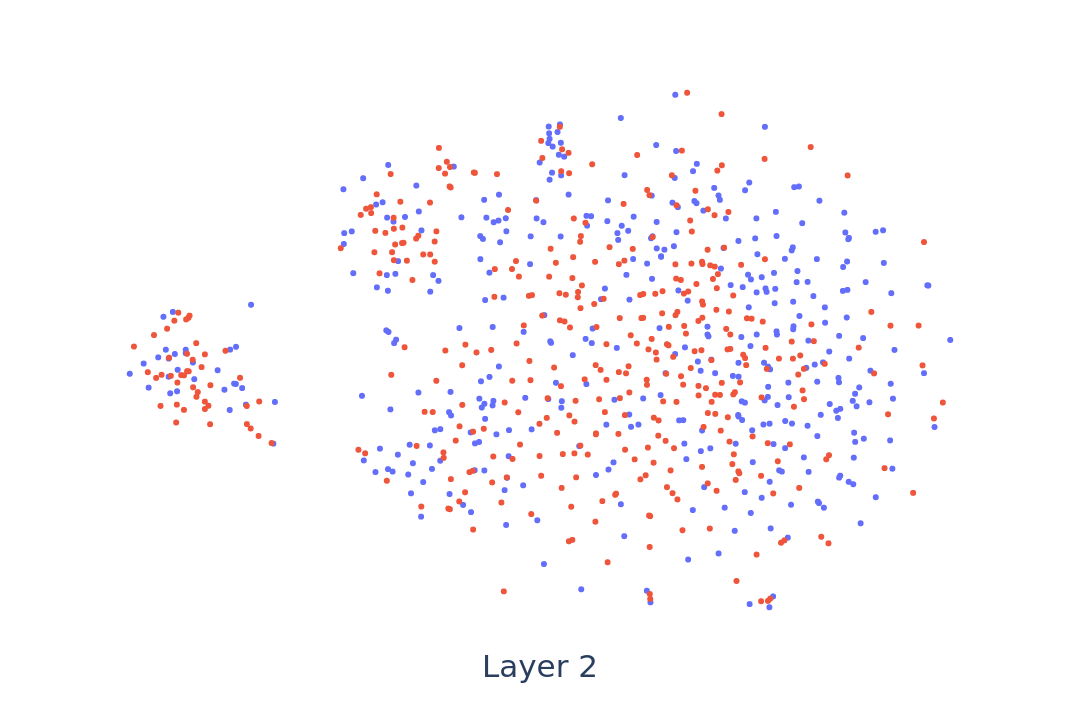}&
     \includegraphics[width=\textwidth]{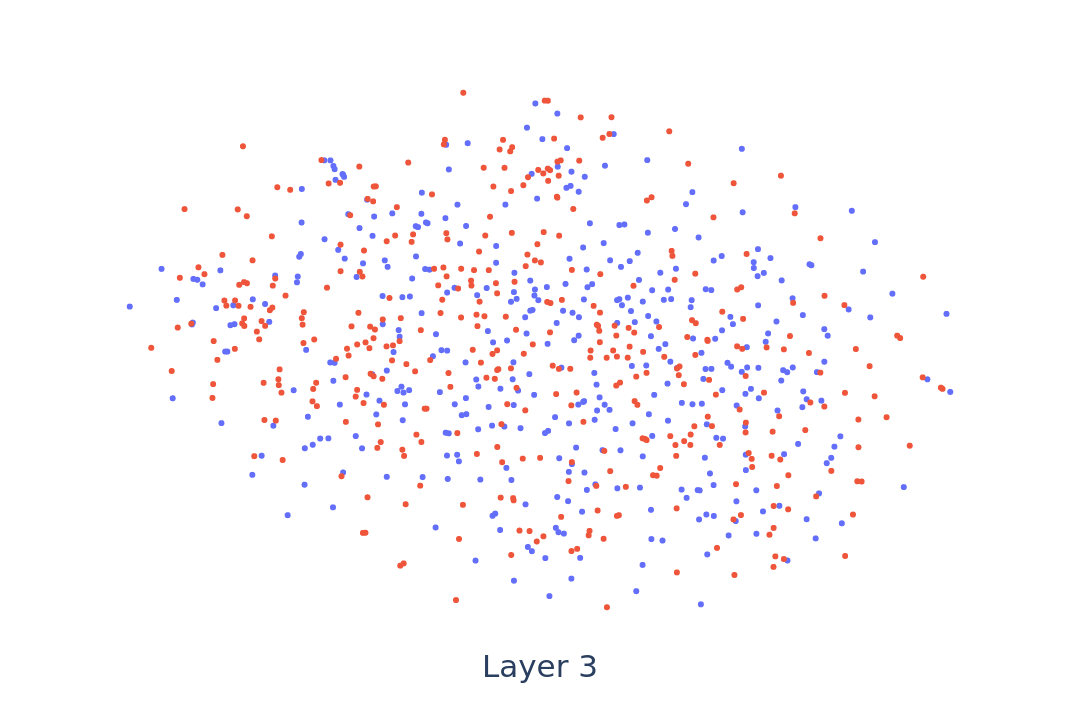}&
     \includegraphics[width=\textwidth]{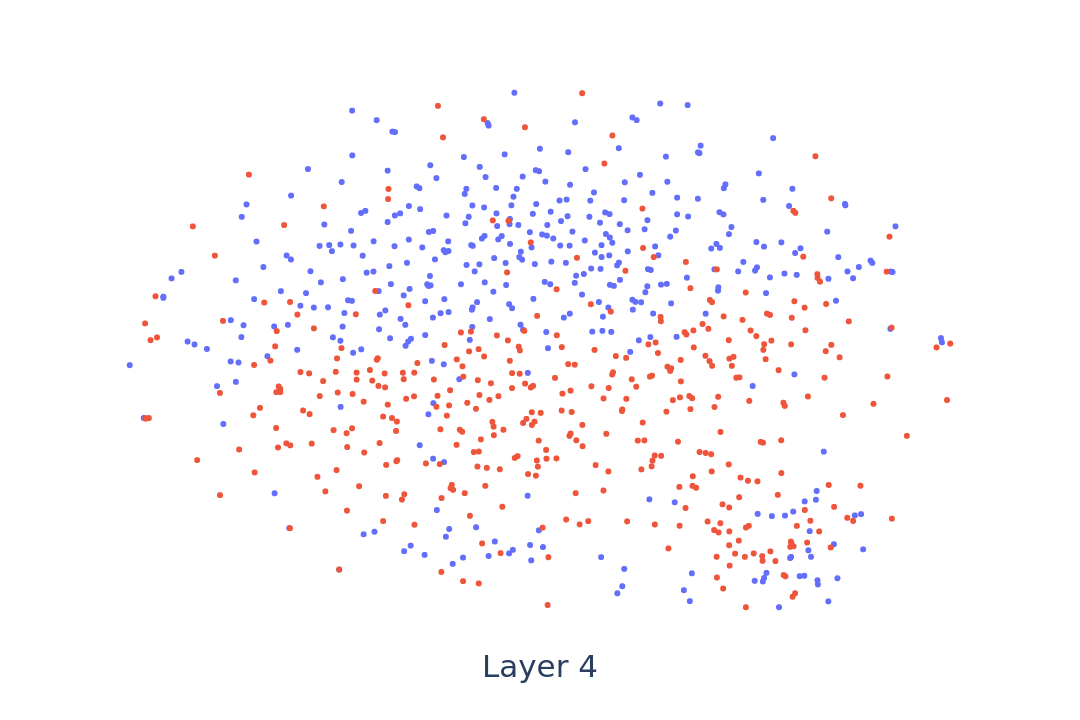} \\
 \includegraphics[width=\textwidth]{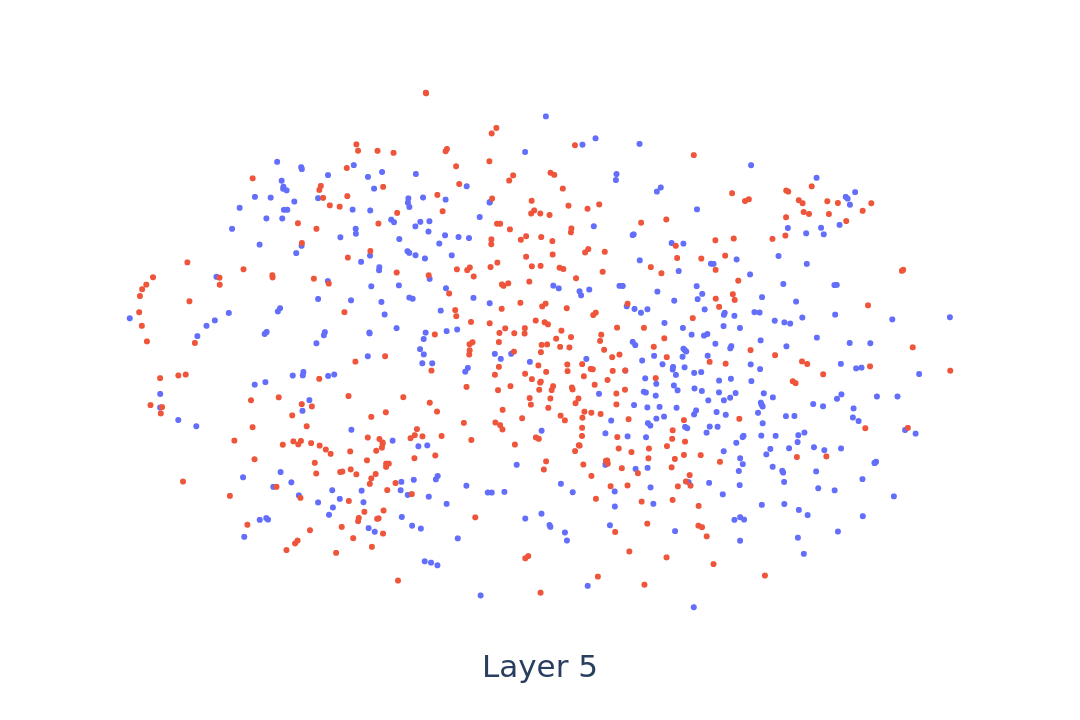}&
     \includegraphics[width=\textwidth]{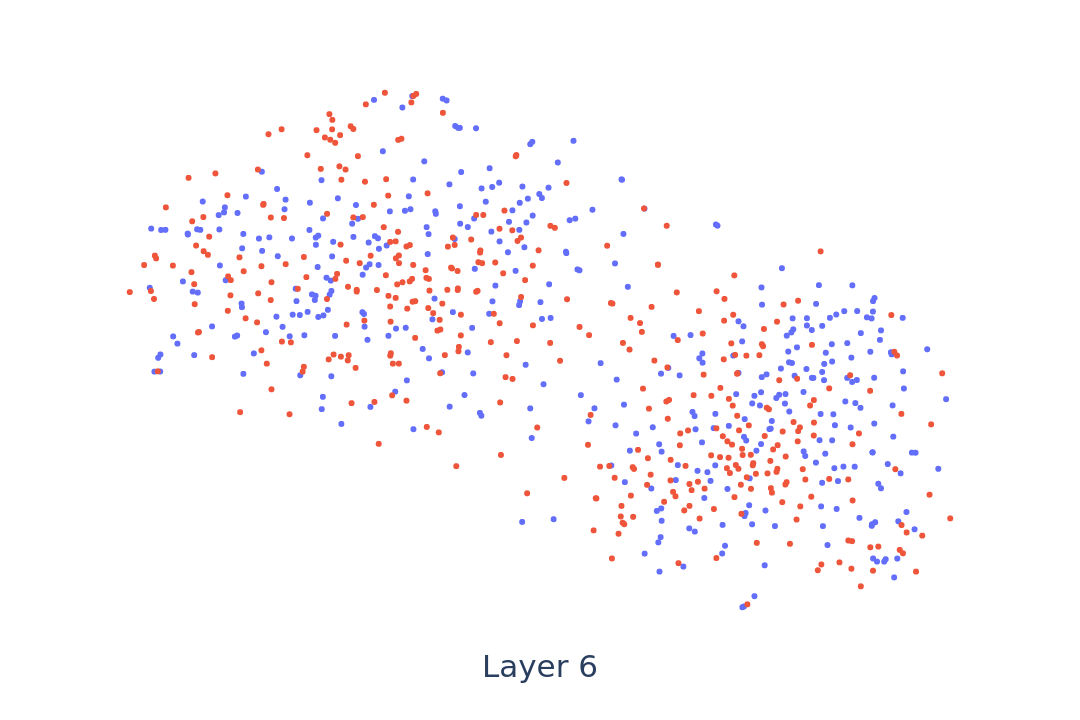}&
     \includegraphics[width=\textwidth]{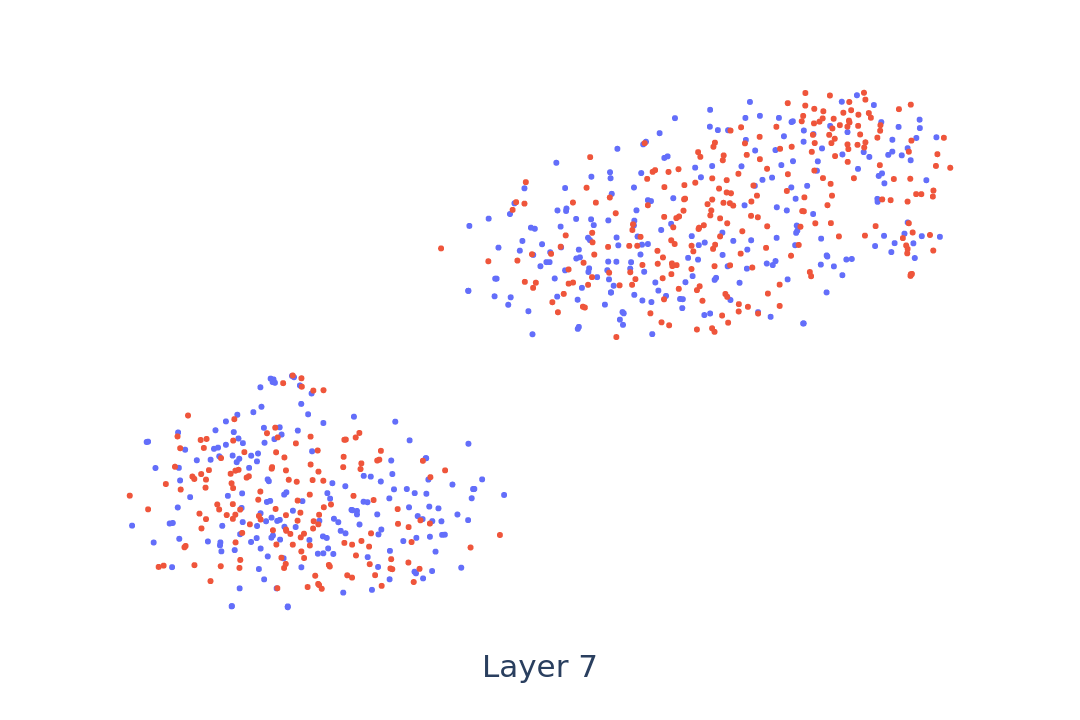}&
     \includegraphics[width=\textwidth]{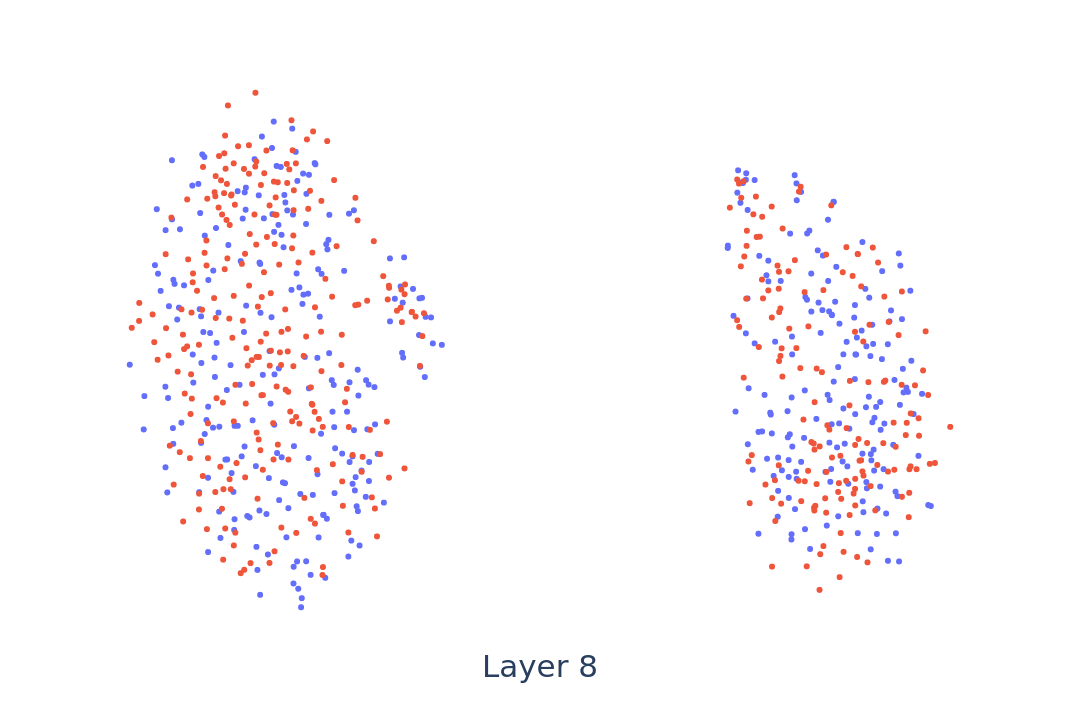} \\
 \includegraphics[width=\textwidth]{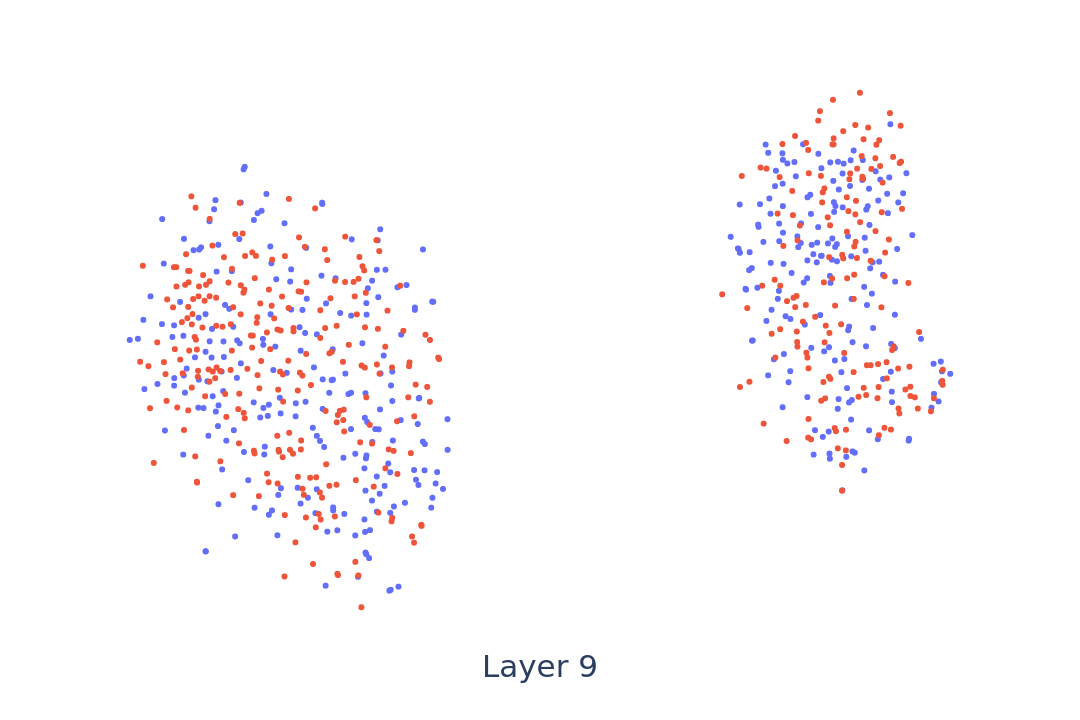}&
     \includegraphics[width=\textwidth]{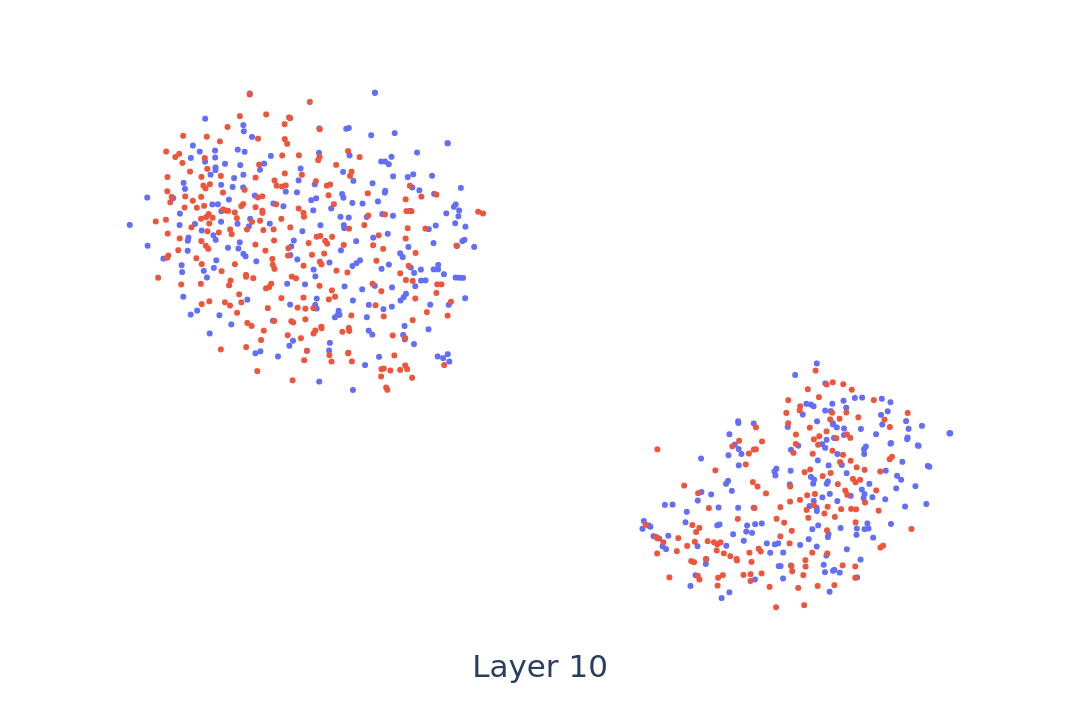}&
     \includegraphics[width=\textwidth]{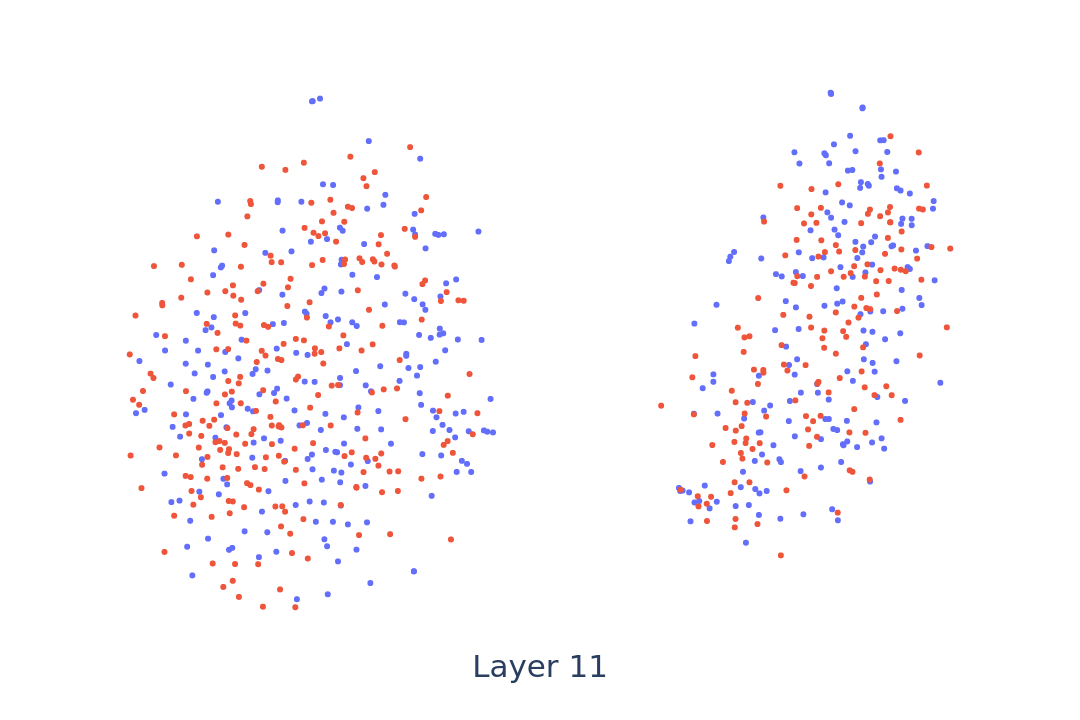}&
     \includegraphics[width=\textwidth]{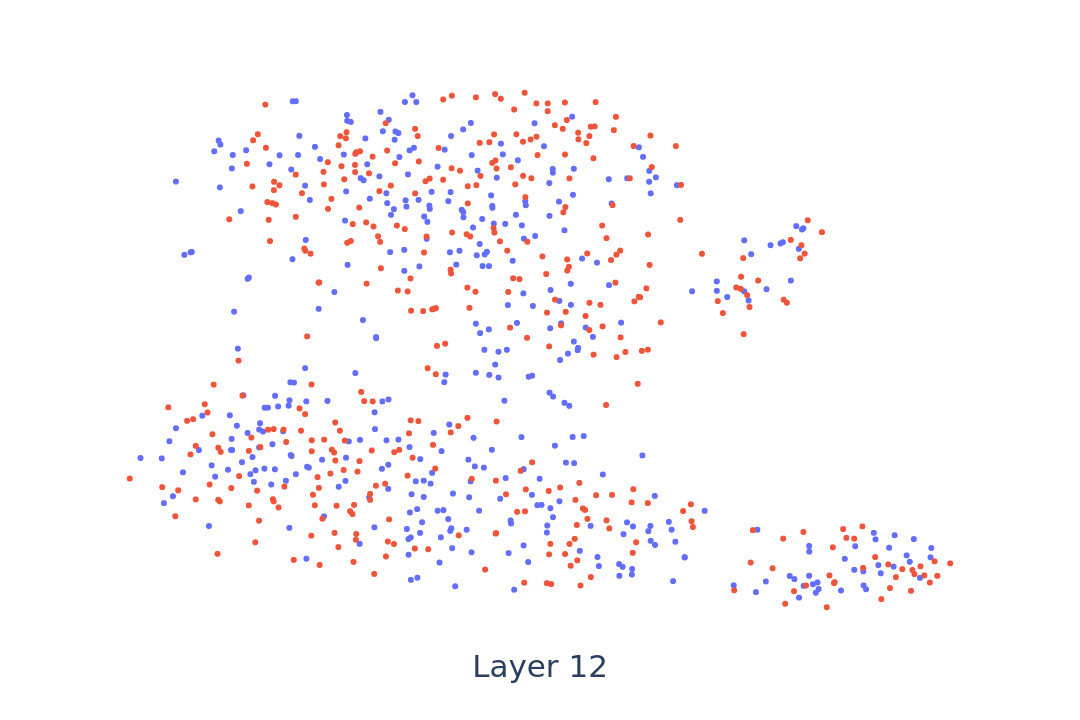} \\
  \end{tabular} %
  }
 \caption{t-SNE plots for the representations from domain adapter trained on \cameraphoto \textrightarrow{} \books{} domain for \amazondataset{}. We reduce divergence between domains for all layers.}
 \label{fig:appendix-amazon-domain-adapter-tsne}
\end{figure*}

\begin{figure*}[t!]
    \centering
    \subfloat[\label{fig:vocab_overlap_amazon}]{\includegraphics[width=0.53\textwidth]{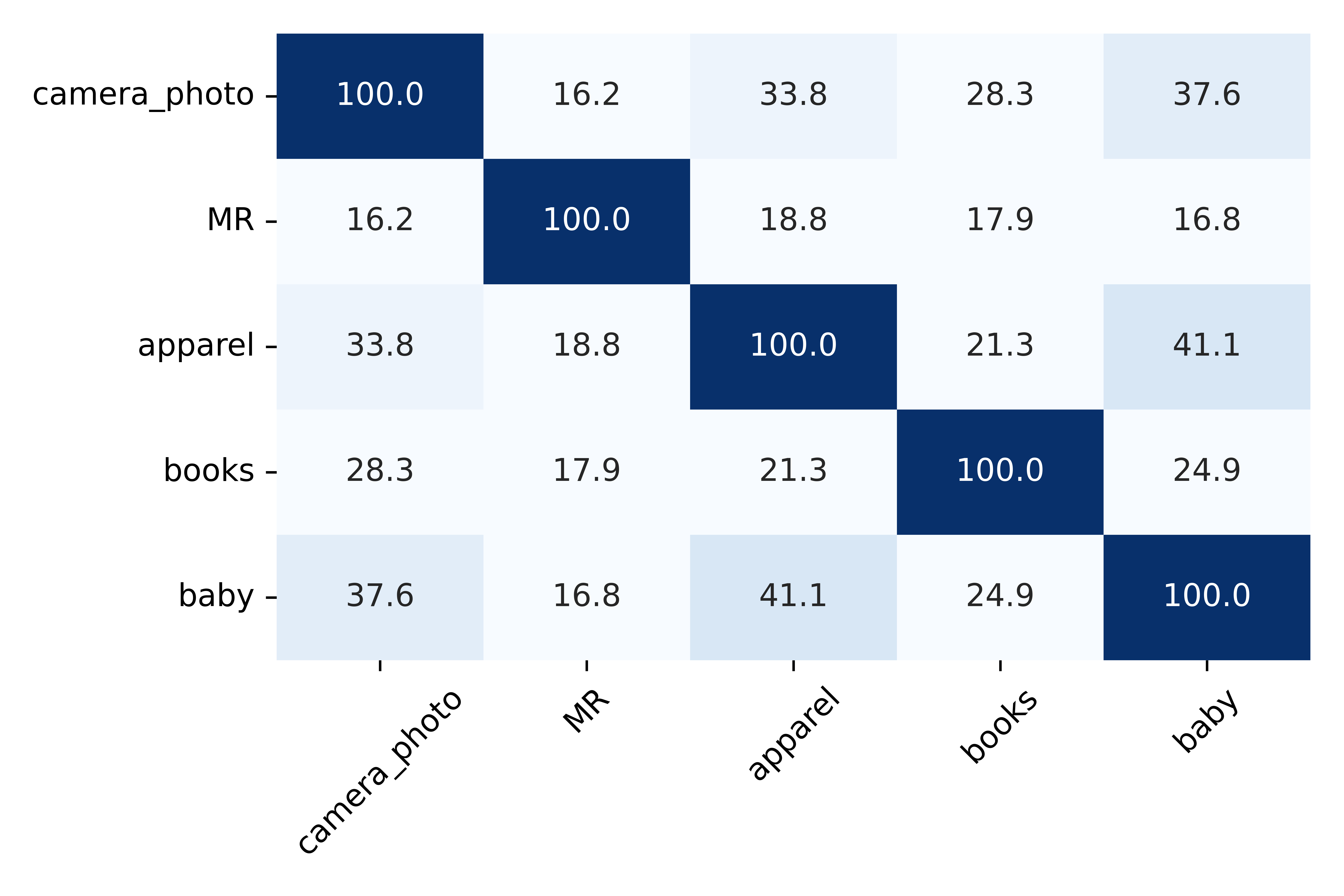}}
    \hspace{0.01cm}
    \subfloat[\label{fig:vocab_overlap_mnli}]{\includegraphics[width=0.53\textwidth]{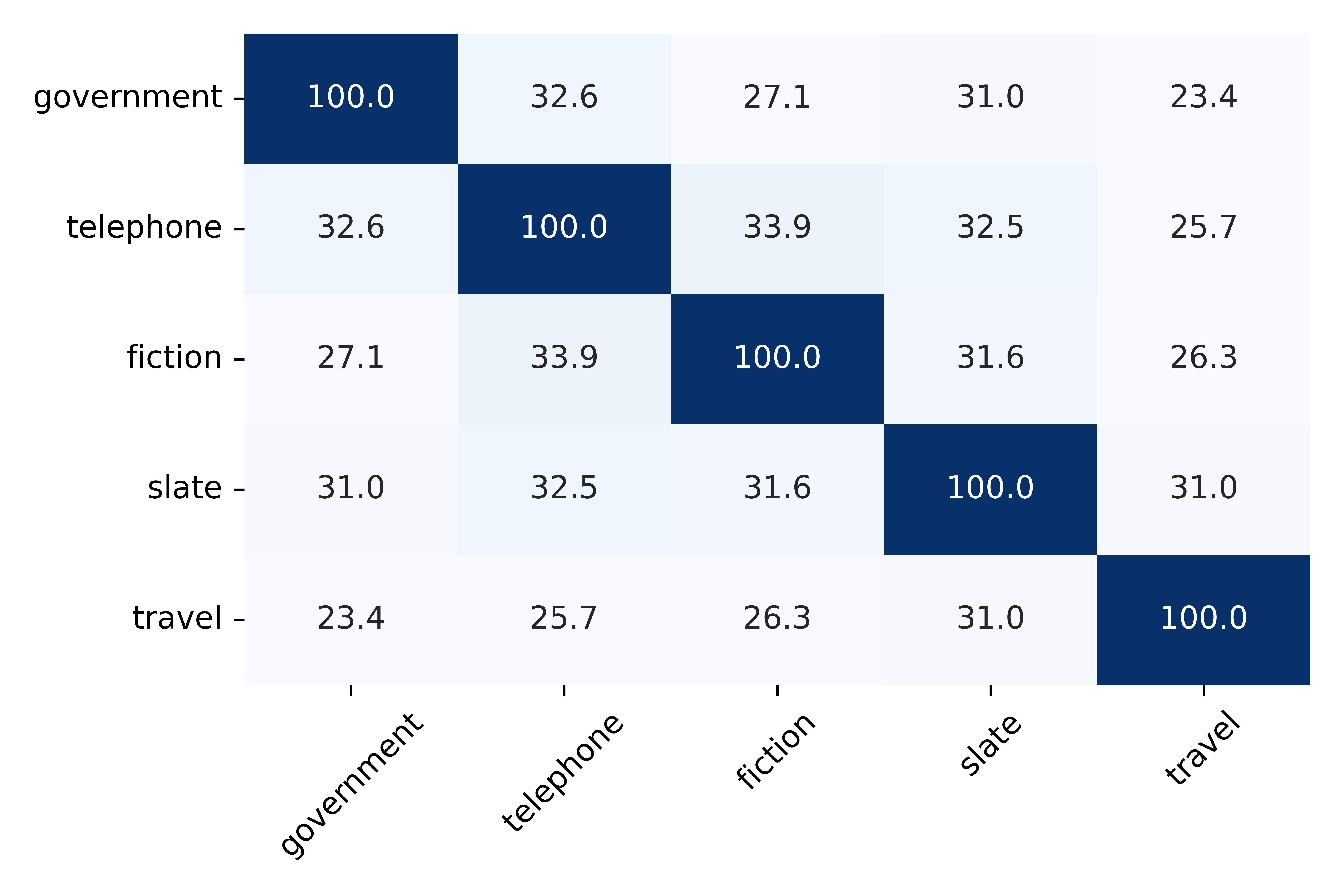}} 
    \caption{(a) Vocabulary overlap (\%) between domains in \amazondataset{}. (b) Vocabulary overlap (\%) between domains in \mnlidataset{}.}
    \label{fig:vocab_overlap}
\end{figure*}

\end{document}